\documentclass[10pt,twocolumn,letterpaper]{article}

\usepackage[pagenumbers]{cvpr} % To force page numbers, e.g. for an arXiv version

\definecolor{cvprblue}{rgb}{0.21,0.49,0.74}
\usepackage[pagebackref,breaklinks,colorlinks,allcolors=cvprblue]{hyperref}
\usepackage{algorithm}
\usepackage{algpseudocode}
\usepackage{listings}
\usepackage[most]{tcolorbox}
\usepackage{xcolor}
\usepackage{booktabs}
\usepackage[table]{xcolor}
\usepackage{tabularx}
\definecolor{sectiongray}{HTML}{F0F0F0}
\definecolor{bestblue}{HTML}{C5E3F6}
\definecolor{secondblue}{HTML}{E6F3FC}
\definecolor{codegreen}{HTML}{418785}
\definecolor{codepink}{HTML}{E72B83}
\definecolor{dropred}{RGB}{180,50,50}
\definecolor{gained}{RGB}{35,135,75}
\definecolor{executionbg}{RGB}{240, 247, 252}
\usepackage{makecell}
\usepackage{cuted}
\usepackage{caption}

\lstdefinestyle{embocode}{
    language=Python,
    basicstyle=\ttfamily\small,
    keywordstyle=\color{black},
    commentstyle=\color{codegreen},
    emph={reset,score,policy,frontier,execute,update},
    emphstyle=\color{codepink},
    numbers=none,
    frame=none,
    showstringspaces=false,
    columns=fullflexible,
    keepspaces=true,
    aboveskip=8pt,
    belowskip=8pt
}

\def\confName{CVPR}
\def\confYear{2027}

\makeatletter
\patchcmd{\@maketitle}
    {\vspace*{24pt}}
    {\vspace*{8pt}}
    {}{}
\makeatother

\title{
    Embo-Harness: Towards Understanding Frontier-Model-as-Policy Harness
    \\[0.2em]
    {\normalfont\normalsize
    \urlstyle{same}
    \url{https://dravenalg.github.io/projects/Embo-Harness/}}
}

\author{
    {\normalsize
    Xiao-Ming Wu\textsuperscript{1},
    Kang Liao\textsuperscript{1},
    Yuexiao Ma\textsuperscript{1},
    Taolin Zhang\textsuperscript{1},
    Mingqian Zhou\textsuperscript{1}}\\[0.1em]
    {\normalsize
    Nuo Chen\textsuperscript{5},
    Ziming Li\textsuperscript{6},
    Yihang Luo\textsuperscript{1},
    Rui Li\textsuperscript{1},
    Zhonghua Wu\textsuperscript{3}}\\[0.1em]
    {\normalsize
    Tianyu Xie\textsuperscript{4},
    Xiawu Zheng\textsuperscript{4},
    Zhongang Cai\textsuperscript{3},
    Zhi Hou\textsuperscript{2},
    Chen Change Loy\textsuperscript{1}\thanks{Corresponding author.}}\\[0.55em]
    {\small
    \textsuperscript{1}S-Lab, Nanyang Technological University
    \quad
    \textsuperscript{2}ACE Robotics
    \quad
    \textsuperscript{3}SenseTime Research}\\[0.15em]
    {\small
    \textsuperscript{4}Xiamen University
    \quad
    \textsuperscript{5}Lenovo Robotics Lab
    \quad
    \textsuperscript{6}University of Connecticut}
}

\begin{document}
\maketitle
\begin{strip}
    \centering
    \vspace{-18mm}
    \includegraphics[width=\linewidth]{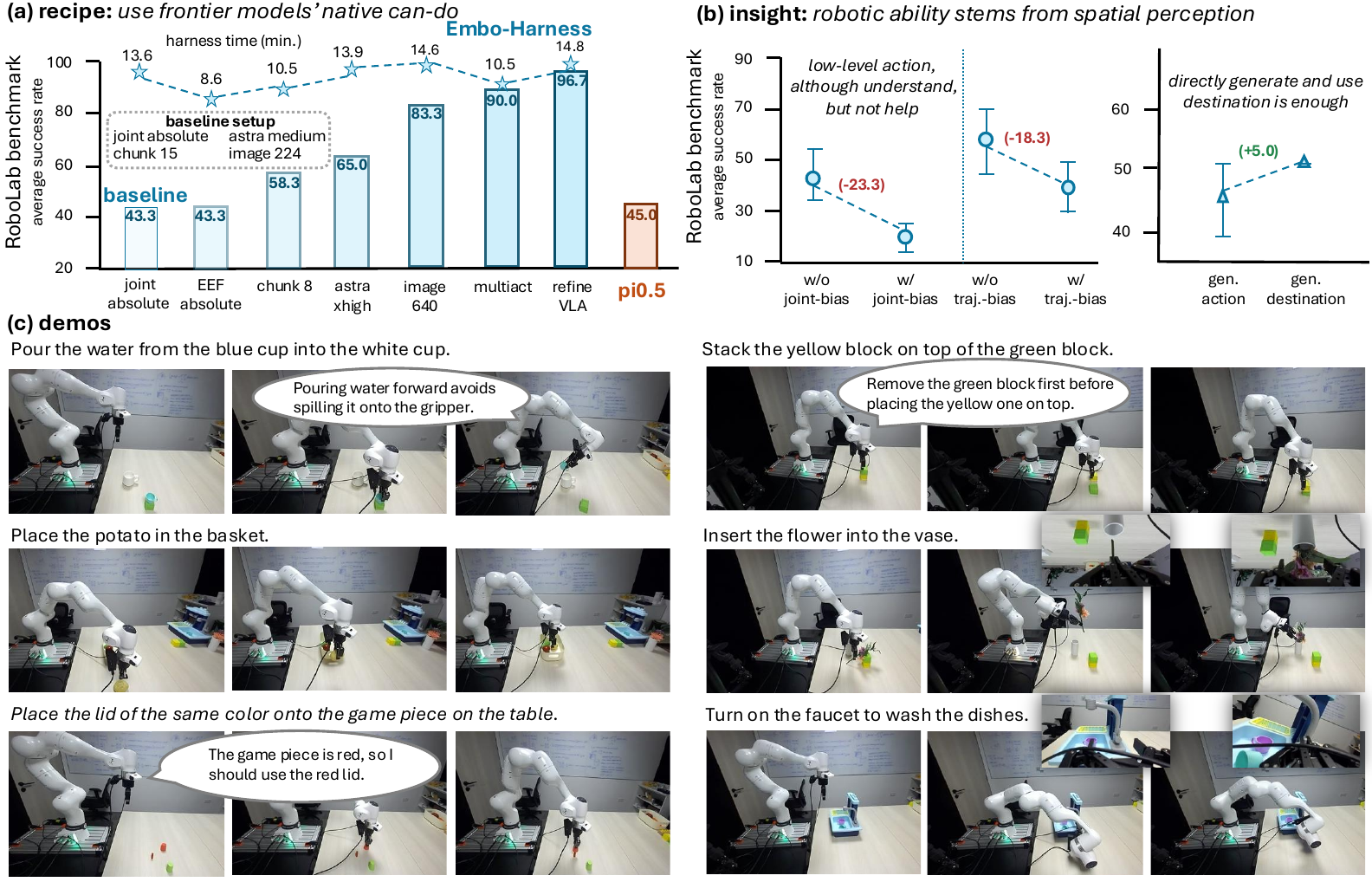}
    \vspace{-8mm}
    \captionsetup{hypcap=false}
    \captionof{figure}{\textbf{Embo-Harness:} \textbf{(a) Recipe.} We progressively develop a frontier-model-as-policy harness on RoboLab benchmark (20 tasks in 3 runs). Short multi-action sequences enable multiple consecutive actions in a single call. \textbf{(b) Insight.} Joint bias refers to direct joint-angle generation without IK assistance, while trajectory bias refers to frame-wise action generation instead of target-pose prediction with interpolation. EEF destination prediction directly generates the final target pose. \textbf{(c) Demos.} Real-world manipulation demos.}
    \captionsetup{hypcap=true}
    \label{fig:teaser}
    \vspace{-2mm}
\end{strip}

\begin{abstract}
Frontier multimodal models exhibit extraordinary generalization in generating robot actions from observations and instructions, enabling a new paradigm in which they serve directly as policies within closed-loop robotic harnesses. However, the design principles of such harnesses, and the capabilities underlying their strong performance, remain unclear. We present Embo-Harness, a framework for systematically studying frontier-model-as-policy harnesses. We investigate four components of the robotic agentic loop, namely, prompting, planning, action execution, and tool use, and distill six empirical findings into a practical recipe. This recipe prioritizes interfaces that exploit the model’s native visual reasoning and tool-use capabilities over handcrafted plans, skills, or action priors, and identifies effective choices for action representation, chunking, reasoning effort, visual observations, short multi-action sequences, and policy refinement. On RoboLab-120, Embo-Harness achieves an 83.3\% success rate, outperforming the strongest reported baseline by 26.2 percentage points. It also succeeds in 15 of 20 zero-shot trials across four real-world manipulation tasks. Beyond performance, controlled interventions indicate that the studied model depends more strongly on accurate spatial target estimation than on explicit low-level joint or trajectory generation. These results provide both a practical recipe for building strong frontier-model-as-policy harnesses and a clearer account of the capabilities that drive their performance.

\end{abstract}    
\section{Introduction}
\label{sec:intro}

Recent advances in frontier models, such as GPT-6-Astra~\cite{2026_gpt6}, are reshaping robotic manipulation~\cite{2026_gpt-6-astra-robodojo, 2026_roborsi}. Their remarkable generalization capabilities enable them to handle complex instructions, diverse environments, long-horizon manipulation, and failure recovery, substantially outperforming existing robot foundation models (RFMs), such as Vision-Language-Action models (VLAs)~\cite{2023_RT-2, 2024_pi0, 2024_OpenVLA, 2025_pi05, 2026_VLANeXt} and World Action Models (WAMs)~\cite{2026_dreamzero, 2026_lingbot-va, 2026_fast-wam}, on recent benchmarks~\cite{2023_LIBERO, 2025_libero-pro, 2026_robolab, 2026_robodojo}, offering a different possibility for future robotic policies.

\begin{figure}[t]
    \centering
    \includegraphics[width=0.85\linewidth]{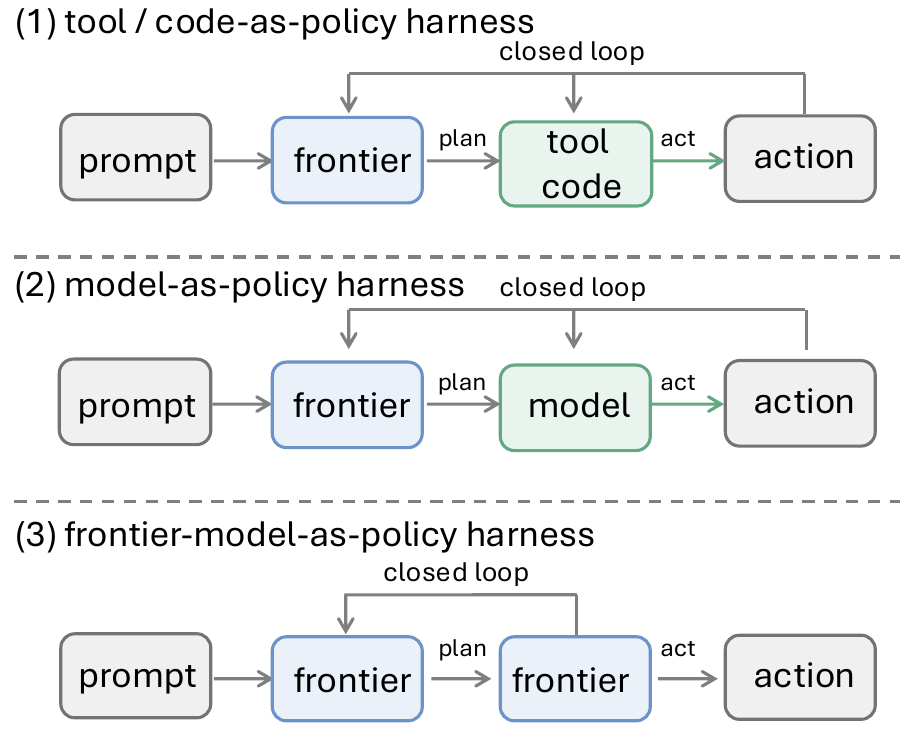}
    \vspace{-2mm}
    \caption{\textbf{Typical Robotic Harness Paradigms.} (1) \textit{Tool/code-as-policy} harnesses leverage frontier models for high-level planning while relying on predefined tools or generated code for action execution. (2) \textit{Model-as-policy} harnesses delegate action generation to specialized policy models. (3) \textit{Frontier-model-as-policy} harnesses directly generate robot actions using frontier models. All three paradigms operate within a closed-loop framework, differing primarily in how robot actions are generated.}
    \vspace{-4mm}
    \label{fig:harness_baselines}
\end{figure}

A robotic harness is the software loop that connects observations and instructions to planning, tool use, action generation, execution, and feedback. As illustrated in Figure~\ref{fig:harness_baselines}, existing robotic harnesses can be broadly divided according to how actions are produced. Tool- or code-as-policy~\cite{2022_zero-shot-planner, 2023_code-as-policies, 2023_voxposer, 2024_ReKep, 2026_dream2flow, 2026_cap-x, 2026_openeta, 2026_RPG, 2026_embodied-turing-machines} harnesses use multimodal models~\cite{2023_qwen, 2024_llama3} for high-level planning but execute actions through predefined tools or generated code. Model-as-policy~\cite{2023_MOO, 2026_harness-vla, 2026_PyRUA-Learn} harnesses instead delegate action generation to specialized learned policies. Recent frontier-model-as-policy harnesses take a further step by using the frontier model itself to plan and generate robot actions within a closed loop. Recent studies have established the feasibility of this paradigm~\cite{2026_show-harness, 2026_gpt-as-policy, 2026_gpt-6-astra-robodojo, 2026_spotter, 2026_physevo, 2026_roborsi}. For example, SenseNova-RoboRSI~\cite{2026_roborsi} improves a frontier-model policy through recursive self-improvement based on prior training datasets and test-time execution rollouts, while GPT-as-Policy~\cite{2026_gpt-as-policy} and related studies~\cite{2026_gpt6, 2026_GPT-Policy} profile the native robotic capabilities of frontier models. However, these works primarily demonstrate or improve the performance of particular systems. They do not systematically study which harness components are responsible for the gains, or why general-purpose frontier models can outperform policies specialized for robotic control.

We address these questions through \textbf{Embo-Harness}, a controlled study for frontier-model-as-policy harnesses. Starting from a common baseline, we vary individual design factors and progressively integrate the effective choices, rather than optimizing the complete system as a single configuration. Our study covers four dimensions of the robotic agentic loop: prompting, planning, action execution, and tool use. The resulting six findings can be summarized into three principles. \textit{First}, suitable action representations and execution strategies are essential: frontier models more effectively generate end-effector actions than joint-space actions and favor short, fixed-length action chunks over adaptive chunking. \textit{Second}, their performance benefits substantially from stronger reasoning and richer visual observations, including higher resolutions and complementary viewpoints. \textit{Third}, tools and learned policies should complement the model’s native capabilities rather than replace them: manually designed plans and failure-derived reminders provide limited benefits, whereas short multi-action sequences and VLA refinement improve effectiveness and efficiency. Guided by this recipe, Embo-Harness achieves an 83.3\% success rate on RoboLab-120, compared with 57.1\% for SenseNova-RoboRSI, the strongest existing baseline. It also completes 15 of 20 zero-shot trials across four real-world manipulation tasks.

Beyond performance, we further investigate why frontier models succeed as robotic policies. Although they can generate joint-level and trajectory-level actions commonly learned by VLAs and WAMs, explicitly requiring such generation degrades performance. In contrast, directly predicting the final EEF destination outperforms short-range action generation, suggesting a stronger reliance on spatial perception than low-level action generation. We further validate this hypothesis through spatial position estimation experiments, observing broadly aligned trends with manipulation performance across reasoning efforts and visual resolutions, as well as comparable errors between direct position estimates and gripper positions in failed grasps, both measured against ground-truth target positions. These findings further highlight the central role of spatial perception in frontier models' robotic capabilities.
\section{Revisiting Agentic Robotics}
\label{sec:revisit}
Robot policies map task instructions and observations to executable actions~\cite{2009_robot_learning_survey, 2013_robot_learning_rl_survey, 2026_robot_learning_vla_survey, 2026_robot_learning_agentic_survey}. Traditionally, these policies relied on handcrafted rules and specialized tools~\cite{1986_layered_control, 2012_OMPL, 2014_moveit, 2023_CuRobo}, while recent advances in neural networks and large-scale robot data have enabled learned policies~\cite{2022_PerAct, 2023_RT-1, 2025_diffusion-policy, 2023_ACT, 2024_Octo} and robot foundation models (RFMs), including Vision-Language-Action models (VLAs)~\cite{2023_RT-2, 2024_pi0, 2024_OpenVLA, 2025_pi05, 2026_VLANeXt} and World Action Models (WAMs)~\cite{2026_dreamzero, 2026_lingbot-va, 2026_fast-wam}.

Beyond standalone policies, \textbf{agentic robotics} employs software-based harnesses to coordinate perception, planning, action execution, and tool use within a closed-loop framework. As illustrated in Fig.~\ref{fig:harness_baselines}, existing robotic harnesses can be broadly categorized into three paradigms based on how robot actions are generated.

\noindent\textbf{Tool/code-as-policy.} 
These harnesses typically leverage vision-language models for perception and high-level planning, while relying on predefined tools or generated code for action execution~\cite{2022_zero-shot-planner, 2023_progprompt, 2023_instruct2act, 2023_code-as-policies, 2023_voxposer, 2024_ReKep, 2024_Statler, 2024_llm-zero-shot-traj, 2026_dream2flow, 2026_cap-x, 2026_openeta, 2026_RPG, 2026_embodied-turing-machines, 2026_thinklocomanip, 2026_mcga, 2026_captain, 2026_AMP1}. Such tools include object detection~\cite{2022_owl-vit}, segmentation~\cite{2023_SAM, 2025_SAM2, 2026_SAM3}, grasp detection~\cite{2023_AnyGrasp, 2024_EconomicGrasp, 2025_motiongrasp}, and motion planning~\cite{2023_CuRobo, 2014_moveit}. Code-as-Policy methods further enable flexible composition and invocation of these tools through code~\cite{2023_code-as-policies, 2026_cap-x, 2026_embodied-turing-machines, 2026_mcga}.

\noindent\textbf{Model-as-policy.}
Rather than relying solely on predefined tools, these harnesses delegate action generation to learned policy models, while VLMs coordinate high-level planning, execution, and feedback~\cite{2022_saycan, 2022_Inner-Monologue, 2023_MOO, 2024_doremi, 2025_Hi-Robot, 2026_harness-vla, 2026_PyRUA-Learn, 2026_astra-ebench, 2026_spatialharness}. By integrating specialized policy models into the agentic loop, these approaches combine high-level reasoning with learned low-level control~.

\noindent\textbf{Frontier-model-as-policy.}
Recent advances in frontier models, such as GPT-6-Astra, have introduced a new paradigm in which frontier models directly generate robot actions, integrating perception, planning, and action generation within a unified agentic loop~\cite{2026_show-harness, 2026_gpt-as-policy, 2026_gpt-6-astra-robodojo, 2026_GPT-Policy, 2026_spotter, 2026_physevo, 2026_bridging-frontier, 2026_roborsi, 2026_fastharness}. Recent studies show that frontier models exhibit remarkable generalization, outperforming several VLA and WAM baselines on robotic manipulation benchmarks~\cite{2026_roborsi, 2026_gpt-6-astra-robodojo}. However, existing studies have primarily focused on demonstrating their capabilities, leaving open how to effectively design frontier-model-as-policy harnesses and what underlying capabilities drive their performance.

Thus, we introduce \textbf{Embo-Harness} to systematically develop a practical harness recipe and investigate the underlying capabilities of frontier models. A concurrent online blog, Astranomy~\cite{2026_astranomy}, similarly explores frontier models for robotic manipulation. Our work further develops a unified recipe through progressive exploration and investigates their underlying capabilities, highlighting the importance of spatial perception over low-level action generation.
\section{Developing Embo-Harness}
\label{sec:method}

\begin{figure}[t]
    \centering
    \includegraphics[width=0.9\linewidth]{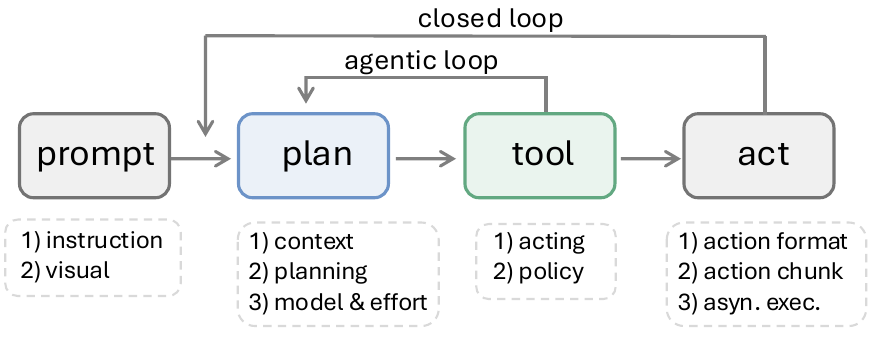}
    \vspace{-2mm}
    \caption{\textbf{Overview of the Robotic Agentic Loop.} A robotic harness consists of four key dimensions: 1) \textit{Prompting}, including task instructions and visual observations; 2) \textit{Planning}, including context management, planning strategies, and model and reasoning effort selection; 3) \textit{Tool Use}, including acting tools and policy models; and 4) \textit{Action Execution}, including action formats, action chunking, and asynchronous execution. These dimensions operate within a closed-loop framework and form the design space systematically explored in Embo-Harness.}
    \vspace{-4mm}
    \label{fig:agentic_loop}
\end{figure}

Frontier models have reshaped robotic harnesses while introducing a fundamental design challenge: how to fully leverage their capabilities within an agentic loop. Toward this end, we systematically explore four key dimensions of the robotic agentic loop: prompting, planning, action execution, and tool use, as illustrated in Fig.~\ref{fig:agentic_loop}. Specifically, prompting covers text prompts and visual inputs; planning involves context management, planning strategies, and model selection; action execution considers action representations, action chunk sizes, and asynchronous inference; and tool use further extends the harness through combined action generation and integration with robot policies. 

\begin{table}[!t]
    \centering
    \small
    \caption{\textbf{Ablation across the robotic harness design space on RoboLab (20 tasks, 3 runs).} We evaluate the effects of different design choices across four key dimensions: action execution, planning, prompting, and tool use. We report success rate (SR), task completion score (Score), and harness execution time.}
    \label{tab:ablation}
    \vspace{-2mm}
    \setlength{\tabcolsep}{2pt}
    \renewcommand{\arraystretch}{0.94}
    \begin{tabular*}{\columnwidth}{@{\extracolsep{\fill}}llcc@{}}
        \toprule
        \textbf{Model} & \textbf{SR (\%)} & \textbf{Score} & \textbf{Harness time (min.)} \\
        \midrule
        \rowcolor{sectiongray}
        \multicolumn{4}{l}{\textbf{Action Execution }} \\
        \midrule
        \rowcolor{sectiongray}
        \multicolumn{4}{l}{\textit{action space}} \\
        joint absolute & 43.3 & 52.2 & 13.6 \\
        joint delta & 48.3 & 53.8 & 12.1 \\
        EEF absolute & 43.3 & 51.9 & 8.6 \\
        EEF delta & 45.0 & 51.4 & 9.0 \\
        \midrule
        \rowcolor{sectiongray}
        \multicolumn{4}{l}{\textit{action chunking}} \\
        chunk 4 & 40.0 & 50.8 & 9.0 \\
        chunk 8 & 58.3 & 68.3 & 10.5 \\
        chunk 15 & 43.3 & 51.9 & 9.0 \\
        chunk 30 & 21.7 & 28.1 & 5.0 \\
        \midrule
        \rowcolor{sectiongray}
        \multicolumn{4}{l}{\textit{action exection}} \\
        syn. exection & 58.3 & 68.3 & 10.5 \\
        asyn. exection & 51.7 & 53.6 & 12.5 \\
        \midrule
        \rowcolor{sectiongray}
        \multicolumn{4}{l}{\textbf{Planning}} \\
        \midrule
        \rowcolor{sectiongray}
        \multicolumn{4}{l}{\textit{model type and effort}} \\
        GPT-6-astra-medium & 58.3 & 68.3 & 10.5 \\
        GPT-6-astra-high & 51.7 & 55.6 & 14.5 \\
        GPT-6-astra-xhigh & 65.0 & 69.2 & 13.9 \\
        GPT-6.1-sol-medium & 40.0 & 44.4 & 12.8 \\
        GPT-6.1-terra-medium & 38.3 & 39.7 & 12.1 \\
        \midrule
        \rowcolor{sectiongray}
        \multicolumn{4}{l}{\textit{planning}} \\
        implicit planning & 65.0 & 69.2 & 13.9 \\
        explicit planning & 61.7 & 65.0 & 16.5 \\
        \midrule
        \rowcolor{sectiongray}
        \multicolumn{4}{l}{\textit{context}} \\
        full context & 65.0 & 69.2 & 13.9 \\
        cut visual history context & \multicolumn{3}{c}{8x token cost (regen. KV-cache)} \\
        \midrule
        \rowcolor{sectiongray}
        \multicolumn{4}{l}{\textbf{Prompting and Visual Inputs}} \\
        \midrule
        \rowcolor{sectiongray}
        \multicolumn{4}{l}{\textit{prompt}} \\
        system prompt only & 65.0 & 69.2 & 13.9 \\
        with notice reminder & 60.0 & 60.3 & 15.3 \\
        \midrule
        \rowcolor{sectiongray}
        \multicolumn{4}{l}{\textit{views}} \\
        two views & 65.0 & 69.2 & 13.9 \\
        single view & 30.0 & 36.4 & 13.1 \\
        \midrule
        \rowcolor{sectiongray}
        \multicolumn{4}{l}{\textit{visual resolution}} \\
        image 224 $\times$ 126  & 65.0 & 69.2 & 13.9 \\
        image 320 $\times$ 180 & 71.7 & 73.6 & 13.4 \\
        image 640 $\times$ 360 & 83.3 & 86.7 & 14.6 \\
        \midrule
        \rowcolor{sectiongray}
        \multicolumn{4}{l}{\textbf{Tool and Policy Integration}} \\
        \midrule
        \rowcolor{sectiongray}
        \multicolumn{4}{l}{\textit{short action sequence as tool}} \\
        one action at a time & 83.3 & 86.7 & 14.6 \\
        with action sequence & 90.0 & 90.0 & 10.5 \\
        \midrule
        \rowcolor{sectiongray}
        \multicolumn{4}{l}{\textit{policy models as tool}} \\
        frontier model only & 90.0 & 90.0 & 10.5 \\
        select policy models & 58.3 & 62.2 & 4.4 \\
        refine policy models & 96.7 & 96.7 & 14.8 \\
        \bottomrule
    \end{tabular*}
    \vspace{-3mm}
\end{table}

\noindent \textbf{Baseline Setup.}
For our exploration, we follow the experimental setup of $\pi_{0.5}$~\cite{2025_pi05}. Specifically, we adopt absolute joint positions as the action representation with an action chunk of 15, providing only task instructions and dual-view images at a resolution of $224 \times 126$ as inputs. We use GPT-6-Astra with medium reasoning effort as our baseline, operating in goal mode for task execution. We adopt a progressive evaluation protocol, where each experiment varies a single design factor while keeping all others fixed. The best-performing configuration identified at each stage is then carried forward to subsequent experiments, progressively refining the harness recipe. The full exploration results are shown in Table~\ref{tab:ablation}. All experiments are conducted on a subset of 20 tasks from the RoboLab benchmark~\cite{2026_robolab}, with $3$ trials per task to reduce the impact of randomness.

\subsection{Action Execution}
\noindent \textbf{Action Format.} 
We first investigate action execution, the dimension most closely related to robot control, by examining whether frontier models can understand and generate different action representations and identifying the most effective one. In robotic manipulation, two common action representations are end-effector (EEF) actions, which specify the spatial position of the robot end-effector, and joint-space actions, which specify the angles of individual robot joints. Given the robot's kinematic parameters, EEF actions can be converted into joint-space actions through inverse kinematics (IK). Both representations can be further expressed as absolute values or relative offsets (delta) from the previous timestep, resulting in four action formats. As shown in Table~\ref{tab:ablation}, \textit{frontier models can handle all four formats with comparable success rates. However, generating joint-space actions is more challenging and incurs longer execution times.} We observe that frontier models often resort to writing code to use IK to compute joint values, directly generating them in only approximately 30\% of cases, typically for simple or repeated actions. Absolute and delta formats exhibit comparable performance and time cost.

\noindent \textbf{Action Chunk.}
We next investigate whether frontier models can adapt their generated actions to different action chunk sizes. We observe that frontier models tend to predict similar target positions regardless of the specified chunk size, suggesting limited awareness of action execution horizons (detail in appendix). Consequently, larger chunks result in slower movements and may exhaust the episode step budget, while smaller chunks may provide insufficient execution time to reach the target, requiring repeated action execution to the same target. As shown in Table~\ref{tab:ablation}, \textit{a chunk size of 8 achieves the best performance by balancing movement speed and execution horizon, outperforming both larger and smaller chunk selections.}

\noindent \textbf{Asynchronous Execution.} 
We next investigate whether frontier models can support asynchronous inference, a technique commonly used in robot policies to reduce execution latency. Specifically, asynchronous inference starts generating the next action chunk before the current chunk finishes executing, allowing the next chunk to be ready upon completion and thereby reducing the idle time between consecutive chunks. For frontier models, this requires anticipating the future robot state after executing the current chunk and generating subsequent actions accordingly. However, Table~\ref{tab:ablation} shows that \textit{frontier models struggle to anticipate such future states and instead focus primarily on immediate movements}. Therefore, we do not employ asynchronous inference in subsequent experiments.

\subsection{Planning}
\noindent \textbf{Model Types.}
Having established the recipe for action execution, we next turn our attention to model selection and planning. We first investigate the impact of different frontier models and reasoning efforts. For instance, GPT models~\cite{2026_gpt6} offer different variants (\eg, 6-Astra, 6.1-Sol, 6.1-Terra) and reasoning efforts (\eg, xhigh, high, and medium). As shown in Table~\ref{tab:ablation}, \textit{stronger model variants and higher reasoning efforts generally achieve better performance.} Interestingly, medium reasoning effort outperforms high, suggesting that more reasoning does not necessarily translate into better action execution.

\noindent \textbf{Explicit Planning.}
We next investigate whether explicit planning improves robotic manipulation performance. Specifically, we compare two strategies: implicit planning, where the goal model directly plans and executes the task, and explicit planning, where an additional model call in plan mode decomposes the task into subtasks, which are then sequentially executed by the goal model. As shown in Table~\ref{tab:ablation}, \textit{explicit planning provides no performance improvement while increasing execution time due to additional planning overhead.}

\noindent \textbf{Context Management.} 
We further investigate whether frontier models require the full interaction history for action generation. During task execution, the interaction history continuously grows with new observations, although earlier observations may not always be necessary for subsequent actions. Our experiments show that \textit{removing historical observations reduces execution time while maintaining comparable performance. However, this substantially decreases the KV-cache hit rate from over 95\% to approximately 25\%, resulting in a 7--8$\times$ increase in uncached token consumption.} 
% Therefore, we retain the full interaction history in subsequent experiments to preserve caching efficiency. %Accordingly, we preserve the full interaction history without truncation.

\subsection{Prompting and Visual Inputs}

\noindent \textbf{Notice Reminder.}
Having explored model selection and planning, we next investigate whether additional prompt guidance improves robotic manipulation performance. Specifically, we employ recursive self-improvement (RSI) to analyze failure cases from baseline rollouts and distill 12 task-agnostic textual reminders, also removing all task-specific information to retain only task-agnostic guidance. As shown in Table~\ref{tab:ablation}, \textit{these reminders decrease performance and also increase execution time, suggesting that introducing failure-derived robotic skills offers limited benefits to frontier models.} Details of reminders are provided in the appendix.
% We therefore exclude these reminders from our final recipe. 

\noindent \textbf{Visual Views.}
Multi-view observations are a key design choice in robotic policies, providing complementary visual information for manipulation. We next investigate whether this design is equally important for frontier-model-as-policy harnesses. Following conventional robot policies, our baseline uses both third-person and wrist-view observations, which we compare against using only a single third-person view. As shown in Table~\ref{tab:ablation}, \textit{removing the wrist view leads to a performance drop, highlighting the importance of complementary visual viewpoints for robotic manipulation.} 
% We therefore retain both views in our final recipe.

\noindent \textbf{Visual Resolution.}
We next investigate the impact of visual input resolution on robotic manipulation performance. Specifically, we evaluate different input resolutions to examine their effects on action execution. As shown in Table~\ref{tab:ablation}, \textit{higher-resolution visual inputs provide finer spatial details, leading to substantial improvements in harness performance.} For the remaining experiments, we adopt a higher visual resolution of $640 \times 360$.

\subsection{Tool and Policy Integration}
\noindent \textbf{Short Multi-action Sequences.}
We further investigate whether additional tools can further enhance harness performance, focusing on short multi-action sequences that enable multiple consecutive actions to be executed together. Specifically, we introduce an additional tool that allows frontier models to generate multiple consecutive actions in a single output without observing intermediate states. As shown in Table~\ref{tab:ablation}, \textit{this strategy further improves both task performance and execution speed.} We further analyze the short multi-action sequences employed during task execution and identify three common patterns: (1) approach, grasp, and retreat; (2) lower, release, and withdraw; and (3) repetition of previously executed action sequences. Overall, short multi-action sequences are used in approximately 30\% of cases, with an average of three actions generated per sequence. These results motivate the inclusion of short multi-action sequences as a tool in our harness.

\noindent \textbf{VLA Integration.}
Finally, we investigate how to integrate VLAs into the harness to assist action execution. Specifically, we explore two strategies: (1) \textit{VLA selection}, where frontier models decide at each step whether to execute actions themselves or delegate execution to a VLA; and (2) \textit{VLA refinement}, where a VLA proposes an action at each step, which frontier models then refine and execute. As shown in Table~\ref{tab:ablation}, VLA selection substantially reduces execution time but also degrades task performance. We observe that frontier models struggle to accurately assess VLA capabilities and tend to delegate most actions to the VLA, even when its predictions are unreliable. In contrast, \textit{VLA refinement achieves the best performance, suggesting that VLAs and frontier models possess complementary capabilities and that refining VLA-proposed actions leads to more effective task execution}. 

It is noteworthy that this result does not rule out selection as a strategy: calibrated exposure to the VLA's successes and failures, or an explicit estimate of its confidence, may enable more reliable delegation. We leave this possibility to future work.

\subsection{Recipe Summary}
\begin{figure}[t]
    \centering
    \includegraphics[width=\linewidth]{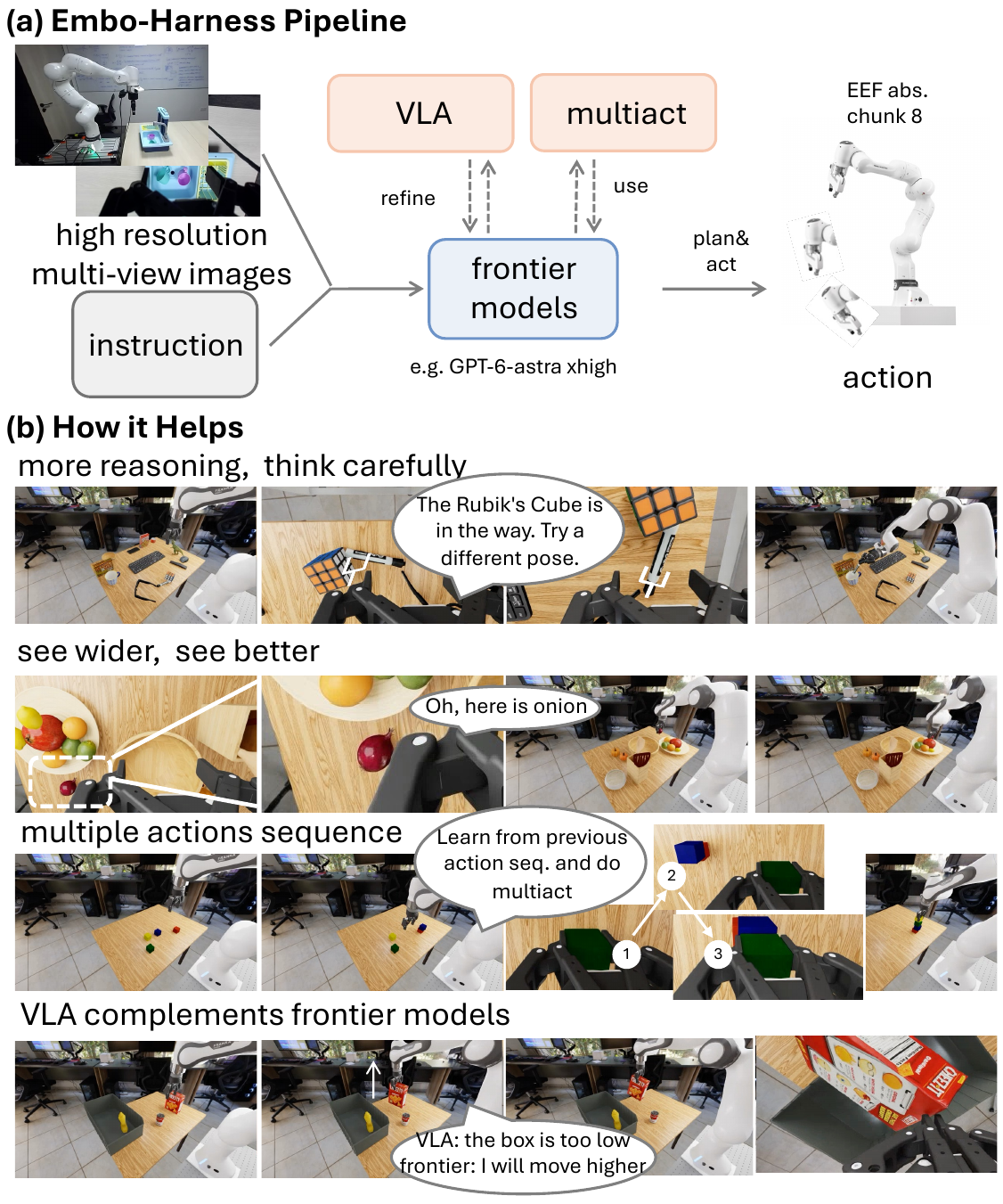}
    \vspace{-8mm}
    \caption{\textbf{Overview of Embo-Harness.} Embo-Harness is designed to leverage the model’s native visual reasoning and tool-use capabilities over handcrafted plans, failure-derived skills, or action priors. The harness takes task instructions and high-resolution visual observations as inputs to GPT-6-Astra with xhigh reasoning effort. The frontier model produces short, fixed-length chunks of absolute end-effector (EEF) actions. Short multi-action sequences and VLA refinement are further integrated as tools to enhance manipulation performance. Zoom in for best view of the demos.}
    \vspace{-2mm}
    \label{fig:embo-harness}
\end{figure}

Through systematic exploration, we distill a practical recipe for building strong frontier-model-as-policy harnesses, as illustrated in Fig.~\ref{fig:embo-harness}. Overall, our findings highlight the importance of leveraging frontier models' native capabilities through effective choices for action representation, chunking, reasoning effort, visual observations, short multi-action sequences, and policy refinement.

\label{subsec:scorer}

\subsection{Evaluating Embo-Harness}
\label{sec:exp}

%\subsubsection{Evaluation Settings}
\noindent \textbf{Benchmark Evaluation.} We evaluate our method on RoboLab-120~\cite{2026_robolab}, a high-fidelity simulation benchmark built on NVIDIA Isaac Lab. The benchmark consists of 120 tabletop manipulation tasks spanning three key competencies: visual, relational, and procedural reasoning. We adopt the DROID robot configuration, featuring a 7-DoF Franka Panda arm equipped with a Robotiq 2F-85 gripper and ZED camera observations. All methods are evaluated out of the box without task-specific fine-tuning, ensuring a fair and consistent comparison. We report two performance metrics, success rate (SR) and task completion score (Score), following Yang~\etal~\cite{2026_robolab}. SR measures the percentage of successfully completed episodes, while Score measures the average proportion of successfully completed subtasks per episode.

\begin{figure}[t]
    \centering
    \includegraphics[width=0.8\linewidth]{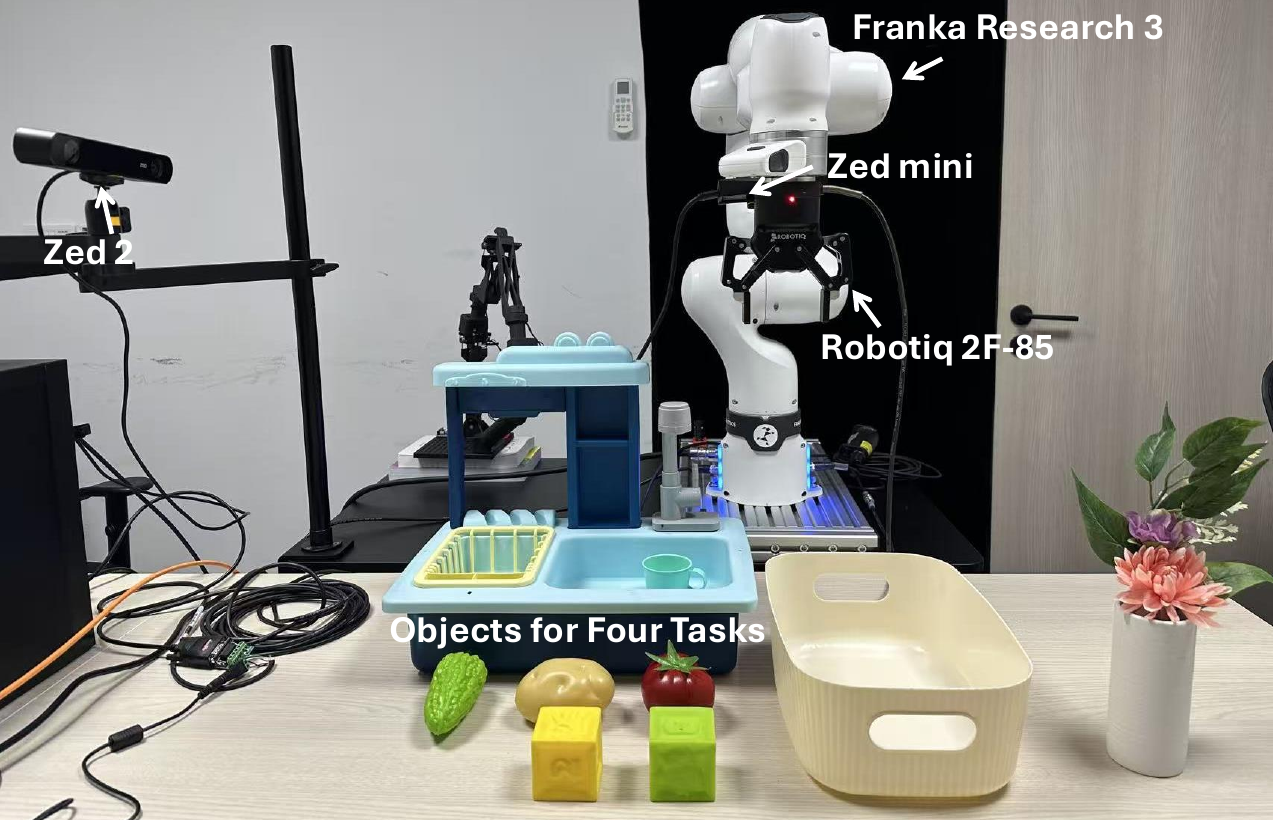}
    \vspace{-2mm}
    \caption{\textbf{Real-world robotic setup for Embo-Harness.} Our setup consists of a Franka Research 3 (FR3) robot arm equipped with a Robotiq 2F-85 gripper, a wrist-mounted ZED Mini camera, and an external ZED 2 camera. We illustrate the objects and workspace used for real-world tabletop manipulation tasks.}
    \vspace{-2mm}
    \label{fig:real_world_setup}
\end{figure}

\noindent \textbf{Real-World Evaluation.} To further validate the effectiveness of our method in real-world environments, we establish a physical robot setup closely matching the RoboLab configuration, as seen in Figure~\ref{fig:real_world_setup}. We evaluate our method on four real-world tabletop manipulation tasks.

\subsubsection{Benchmark Evaluation Results}

We evaluate Embo-Harness on the full RoboLab-120 benchmark against three categories of baselines: VLAs, WAMs, and agentic methods. Among the agentic baselines, SenseNova-RoboRSI~\cite{2026_roborsi} is a closely related frontier-model-as-policy approach that, like ours, employs GPT-6-Astra for robot action generation. As shown in Table~\ref{tab:robolab}, Embo-Harness achieves an SR of 83.3\% and a Score of 89.2, outperforming all compared methods on both metrics where reported. Notably, relative to SenseNova-RoboRSI, the strongest baseline, it improves SR by 26.2 percentage points and Score by 19.3. These results demonstrate the effectiveness of our systematically developed harness recipe, highlighting the importance of leveraging frontier models' native capabilities rather than imposing handcrafted robotic priors. We note that the concurrent work AMP1~\cite{2026_AMP1} also reports strong performance. However, its reliance on an external high-accuracy depth estimator makes the evaluation settings different, precluding a direct comparison.

\begin{table}[t]
    \centering
    \small
    \caption{\textbf{Comparison on the full RoboLab-120 benchmark.}
    We report the success rate (SR) and score.
    The \colorbox{bestblue}{\textbf{best}} and \colorbox{secondblue}{\underline{second-best}} results are highlighted.}
    \vspace{-2mm}
    \label{tab:robolab}
    \setlength{\tabcolsep}{2pt}
    \renewcommand{\arraystretch}{1.05}
    \begin{tabular*}{\columnwidth}{@{\extracolsep{\fill}}llcc@{}}
        \toprule
        \textbf{Policy} & \textbf{Type} & \textbf{SR (\%)} & \textbf{Score} \\
        \midrule
        \rowcolor{sectiongray}
        \multicolumn{4}{l}{\textit{agentic methods}} \\
        \textbf{Embo-Harness} & Agentic & \cellcolor{bestblue}\textbf{83.3} & \cellcolor{bestblue}\textbf{89.2} \\
        SenseNova-RoboRSI~\cite{2026_roborsi}
            & Agentic & \cellcolor{secondblue}\underline{57.1} & \cellcolor{secondblue}\underline{69.9} \\
        VoLo~\cite{2026_volo}
            & Agentic & 28.2 & 44.1 \\
        \midrule
        \rowcolor{sectiongray}
        \multicolumn{4}{l}{\textit{policy models}} \\
        FLUX 3 Action~\cite{2026_flux-3-action}
            & WAM & 42.9 & -- \\
        Atomic-WAM~\cite{2026_atomic-wam}
            & VLM+WAM & 39.6 & 55.5 \\
        Cosmos3-Nano-Policy~\cite{2026_cosmos3}
            & WAM & 36.8 & 51.9 \\
        $\pi_{0.5}$~\cite{2025_pi05}
            & VLA & 28.0 & 43.4 \\
        DreamZero~\cite{2026_dreamzero}
            & WAM & 25.7 & 39.8 \\
        Cosmos3-Edge-Policy~\cite{2026_cosmos3}
            & WAM & 22.9 & -- \\
        $\pi_0$-FAST~\cite{2025_fast}
            & VLA & 15.5 & 26.9 \\
        GR00T N1.6~\cite{2025_gr00t}
            & VLA & 7.2 & 17.1 \\
        $\pi_0$~\cite{2024_pi0}
            & VLA & 5.0 & 12.2 \\
        paligemma-binning~\cite{2024_paligemma}
            & VLA & 3.4 & 9.9 \\
        \bottomrule
    \end{tabular*}
    \vspace{-4mm}
\end{table}

\begin{table}[t]
    \centering
    \small
    \caption{\textbf{Real-world evaluation of Embo-Harness.} We compare Embo-Harness with the baseline across four real-world manipulation tasks: block stacking, fruit packing, dish washing, and flower arranging. We report the success/total number for each task.}
    \vspace{-2mm}
    \label{tab:real_world}
    \setlength{\tabcolsep}{2pt}
    \renewcommand{\arraystretch}{1.05}
    \begin{tabular*}{\columnwidth}{@{\extracolsep{\fill}}lcc@{}}
        \toprule
        \textbf{Policy} & \textbf{Block Stacking} & \textbf{Fruit Packing} \\
        \midrule
        Baseline & 4/5 & 3/5 \\
        Embo-Harness & \textbf{5/5} & \textbf{5/5} \\
        \midrule
        \textbf{Policy} & \textbf{Dish Washing} & \textbf{Flower Arranging} \\
        \midrule
        Baseline & 0/5 & 0/5 \\
        Embo-Harness & \textbf{2/5} & \textbf{3/5} \\
        \bottomrule
    \end{tabular*}
    \vspace{-4mm}
\end{table}

\subsubsection{Real-World Evaluation Results}

We further evaluate Embo-Harness on four real-world manipulation tasks: block stacking, fruit packing, dish washing, and flower arranging. All experiments are conducted in a purely zero-shot setting, without any task-specific memory or hints. As shown in Table~\ref{tab:real_world}, Embo-Harness consistently outperforms our baseline setup, achieving 5/5 successes in both block stacking and fruit packing, 2/5 in dish washing, and 3/5 in flower arranging. The remaining failures primarily stem from limitations in fine-grained spatial perception. Specifically, dish washing requires precise localization of a small faucet switch, which remains challenging for current frontier models. Similarly, flower arranging requires accurate alignment with the vase opening, which becomes difficult to observe from the wrist-view camera at tilted orientations. Despite these challenges, Embo-Harness demonstrates promising zero-shot generalization to real-world robotic manipulation.
\section{Why Frontier Models Succeed in Robotics}
\label{sec:why_succeed}
The emergence of frontier models as competitive robot policies raises a broader question for embodied intelligence: what capabilities enable general-purpose multimodal models, not specifically designed for low-level robot control, to rival or even surpass specialized VLA and WAM policies? Answering this question is important not only for understanding current systems, but also for determining which capabilities and training signals future robot foundation models should prioritize. We investigate three related questions: whether frontier models encode low-level joint and trajectory patterns, whether such knowledge contributes to successful manipulation, and, if not, which more general capability accounts for their performance.

\subsection{Evaluating Low-level Action Generation}
To answer the first two questions, we investigate whether frontier models exhibit characteristic biases of low-level robotic actions and whether these biases contribute to their manipulation capabilities. Specifically, we consider two levels of low-level actions: 1) \textbf{trajectory-level bias}, which captures frame-to-frame motion patterns, and 2) \textbf{joint-level bias}, which captures the movements of individual robot joints. Existing VLAs and WAMs typically learn low-level action patterns through imitation learning, with most methods modeling trajectory-level patterns with an action chunking strategy and some, such as $\pi_{0.5}$~\cite{2025_pi05}, additionally modeling joint-level patterns. We therefore examine whether frontier models exhibit these two biases and whether explicitly leveraging them improves manipulation performance.

\noindent\textbf{Joint-level bias.} We compare direct joint-angle generation (disable inverse kinematics (IK) usage) with joint-angle generation assisted by inverse kinematics (IK). As shown in Table~\ref{tab:low_level_action}, direct joint control achieves a 20.0\% zero-shot success rate, indicating that frontier models possess some understanding of joint-level actions. However, this performance is 23.3 percentage points lower than that of IK-assisted control, suggesting that explicitly generating joint-level actions does not improve manipulation performance.

\noindent\textbf{Trajectory-level bias.} We further compare explicit multi-frame action generation, where frontier models predict an action for each frame, with predicting a future target pose several frames ahead. As shown in Table~\ref{tab:low_level_action}, explicit multi-frame action generation (indicated by the row `direct chunk 8') achieves a 40.0\% zero-shot success rate, demonstrating that frontier models can generate low-level trajectories. But this approach underperforms target pose prediction by 18.3 percentage points (indicated by the row `chunk 8'), suggesting that explicitly modeling frame-level motion patterns also fails to improve manipulation performance.

Overall, these results show that frontier models possess some ability to generate joint-level actions and frame-level trajectories. However, the evaluated harness performs substantially better when the model predicts spatial targets and delegates low-level execution to IK or trajectory interpolation. This suggests that its strong manipulation performance does not primarily depend on explicit low-level action generation, although the model may still use implicit action knowledge during spatial reasoning and target selection. We further find that frontier models generate similar action displacements across different execution horizons, indicating limited sensitivity to frame-level temporal information. Their rollout trajectories also exhibit broader spatial distributions than those of $\pi_{0.5}$~\cite{2025_pi05}, suggesting distinct motion patterns from policies trained on low-level actions. These findings provide additional evidence for our conclusions, with detailed results presented in the supplementary material.

\begin{table}[t]
    \centering
    \small
    \caption{\textbf{Probing robot-specific action biases in GPT-Astra on RoboLab (20 tasks in 3 runs).} We investigate two potential action biases: direct joint control versus IK/FK-assisted control, and direct action chunk generation versus end-effector prediction followed by interpolation. While GPT-Astra demonstrates some ability to generate robot actions directly, the substantial performance drops suggest that its strong manipulation capabilities are not primarily driven by learned robot action priors. We report success rate (SR) and task completion score (Score).}
    \vspace{-2mm}
    \label{tab:low_level_action}
    \setlength{\tabcolsep}{2pt}
    \renewcommand{\arraystretch}{1.05}
    \begin{tabular*}{\columnwidth}{@{\extracolsep{\fill}}lcc@{}}
        \toprule
        \textbf{Policy} & \textbf{SR (\%)} & \textbf{Score} \\
        \midrule
        \rowcolor{sectiongray}
        \multicolumn{3}{l}{\textit{joint bias}} \\
        joint absolute with IK/FK & 43.3 & 52.2 \\
        directly joint absolute &
        20.0 {\scriptsize\textcolor{dropred}{(-23.3)}} &
        27.2 {\scriptsize\textcolor{dropred}{(-25.0)}} \\
        \midrule
        joint delta with IK/FK & 48.3 & 53.8 \\
        directly joint delta &
        16.7 {\scriptsize\textcolor{dropred}{(-31.6)}} &
        26.6 {\scriptsize\textcolor{dropred}{(-27.2)}} \\
        \midrule
        \rowcolor{sectiongray}
        \multicolumn{3}{l}{\textit{action chunk bias}} \\
        chunk 8 & 58.3 & 68.3 \\
        direct chunk 8 & 40.0 {\scriptsize\textcolor{dropred}{(-18.3)}} & 43.6 {\scriptsize\textcolor{dropred}{(-24.7)}} \\
        \midrule
        chunk 15 & 43.3 & 51.9 \\
        direct chunk 15 & 33.3 {\scriptsize\textcolor{dropred}{(-10.0)}} & 36.7 {\scriptsize\textcolor{dropred}{(-15.2)}} \\
        \bottomrule
    \end{tabular*}
    \vspace{-2mm}
\end{table}

\begin{table}[t]
    \centering
    \small
    \caption{\textbf{Probing spatial perception in GPT through EEF control on RoboLab (20 tasks in 3 runs).} We compare absolute and delta end-effector control with direct EEF destination prediction to investigate the role of spatial perception in robotic manipulation. We report success rate (SR) and task completion score (Score).}
    \vspace{-2mm}
    \label{tab:spatial_perception}
    \setlength{\tabcolsep}{2pt}
    \renewcommand{\arraystretch}{1.05}
    \begin{tabular*}{\columnwidth}{@{\extracolsep{\fill}}lcc@{}}
        \toprule
        \textbf{Policy} & \textbf{SR (\%)} & \textbf{Score} \\
        \midrule
        EEF absolute & 43.3 & 51.9 \\
        EEF delta & 45.0 & 51.4 \\
        EEF destination & 50.0 {\scriptsize\textcolor{gained}{(+5.0)}} & 55.0 {\scriptsize\textcolor{gained}{(+3.1)}} \\
        \bottomrule
    \end{tabular*}
    \vspace{-4mm}
\end{table}

\subsection{Evaluating Spatial Perception}
We now turn to the third question: if low-level action understanding is not the primary driver of frontier models' manipulation capabilities, where do these capabilities come from? We hypothesize that their strength primarily stems from accurate 3D spatial perception. To investigate this hypothesis, we compare two action generation strategies: generating short-range action chunks, as in our previous experiments, and directly predicting the final EEF destination. As shown in Table~\ref{tab:spatial_perception}, directly predicting the final destination achieves slightly higher success rates than short-range action generation, suggesting that spatial perception may play a more fundamental role than low-level action generation.

To further examine this hypothesis, we directly evaluate the 3D position estimation accuracy of frontier models under different reasoning efforts, visual resolutions, and movement priors. As shown in Table~\ref{tab:position_estimation}, higher reasoning effort and visual resolution substantially reduce spatial position estimation errors, broadly aligning with the manipulation performance trends in Table~\ref{tab:ablation}. These results suggest that improved spatial perception may contribute to stronger manipulation capabilities. Interestingly, the position estimation errors are comparable in magnitude to the gripper-to-target errors observed in failed grasps under short-range action generation (11.04 cm on average), suggesting that manipulation accuracy is closely related to spatial position estimation. 
We further observe overlapping trajectory distributions between the two strategies, suggesting shared spatial characteristics. These analyses, detailed in the supplementary material, further support the important role of spatial perception in frontier models' robotic capabilities.

\begin{table}[t]
    \centering
    \small
    \caption{\textbf{Evaluation of spatial position estimation and grasp execution errors (20 tasks, 1 run).}
    We report average and median position errors (cm), along with the percentage of errors within 5 cm, under different reasoning efforts, visual resolutions, and movement priors. We further compare these results with the gripper-to-target errors observed in failed grasp executions.}
    \vspace{-2mm}
    \label{tab:position_estimation}
    \setlength{\tabcolsep}{2pt}
    \renewcommand{\arraystretch}{1.05}
    \begin{tabularx}{\columnwidth}{@{}Xccc@{}}
        \toprule
        \textbf{Setting} & \textbf{avg. (cm)} & \textbf{med. (cm)} & \textbf{$\leq$5 cm (\%)} \\
        \midrule
        \rowcolor{sectiongray}
        \multicolumn{4}{l}{\textit{model effort}} \\
        medium & 14.10 & 11.24 & 7.4 \\
        high   & 10.67 & 9.21 & 10.0 \\
        xhigh  & 9.07 & 7.04 & 28.1 \\
        \midrule
        \rowcolor{sectiongray}
        \multicolumn{4}{l}{\textit{visual resolution}} \\
        224 $\times$ 126 & 14.14 & 11.39 & 10.0 \\
        320 $\times$ 180 & 11.57 & 9.58 & 12.5 \\
        640 $\times$ 360 & 9.07 & 7.04 & 28.1 \\
        \midrule
        \rowcolor{sectiongray}
        \multicolumn{4}{l}{\textit{moving prior}} \\
        normal & 9.07 & 7.04 & 28.1 \\
        moving closer & 7.15 & 5.48 & 48.4 \\
        \midrule
        \rowcolor{executionbg}
        \multicolumn{4}{l}{\textit{execution reference (failed grasps)}} \\
        \rowcolor{executionbg}
        grasp position error & 11.04 & 8.39 & 17.2 \\
        \bottomrule
    \end{tabularx}
    \vspace{-4mm}
\end{table}

\subsection{Discussion}
Frontier models are most effective when used to infer task-relevant spatial targets, rather than low-level actions. Three observations from our experiments support this view. First, direct joint and frame-level trajectory generation is possible but consistently underperforms IK-assisted or target-pose control. Second, stronger reasoning and higher visual resolution reduce 3D localization errors in parallel with improvements in task success. Third, real-world failures are concentrated in tasks requiring precise localization and alignment. These findings do not imply that frontier models lack implicit action knowledge or that low-level action data are unnecessary, as the observed performance gaps may also reflect limitations in action interfaces and numerical generation. Nevertheless, they highlight the importance of spatial perception and motivate greater emphasis on spatially grounded models, datasets, and benchmarks. More broadly, these insights suggest a promising direction for developing new robot policy paradigms beyond conventional VLAs and WAMs, potentially achieving strong manipulation performance and efficient execution without relying heavily on low-level action imitation.

\section{Conclusion}
\label{sec:conclusion}
We have introduced Embo-Harness to investigate how to build strong frontier-model-as-policy harnesses and why frontier models achieve remarkable robotic manipulation performance. Through a systematic exploration of prompting, planning, action execution, and tool use, we distilled six key findings into a practical recipe that emphasizes leveraging frontier models' native capabilities rather than imposing handcrafted plans, skills, or action priors. Guided by this recipe, Embo-Harness achieves state-of-the-art performance on RoboLab-120 and demonstrates strong performance in real-world experiments. Beyond harness design, we further investigated the underlying capabilities of frontier models, revealing that their manipulation performance primarily stems from accurate spatial perception rather than low-level action understanding. These findings not only provide practical guidance for developing stronger robotic harnesses but also suggest a promising direction for future robot datasets and foundation models.

\noindent \textbf{Limitations and Future Research.}
Our findings depend on the capabilities of current frontier models and may evolve with future advances. While spatial perception may enable general-purpose gripper-based manipulation, dexterous manipulation, locomotion, and tactile perception require additional low-level and dynamic capabilities, presenting promising directions for future research.

{
    \small
    \bibliographystyle{ieeenat_fullname}
    \bibliography{main}

@String(ICCV= {Int. Conf. Comput. Vis.})

@String(ECCV= {Eur. Conf. Comput. Vis.})

@String(ICLR = {Int. Conf. Learn. Represent.})

@String(ICCV  = {ICCV})

@String(ECCV  = {ECCV})

@String(ICLR  = {ICLR})

@String(ICRA   = {ICRA})

@String(IROS   = {IROS})

@String(CORL   = {CoRL})

@String(RSS    = {RSS})

@String(IJRR   = {IJRR})

@String(TRO    = {IEEE T-RO})

@String(RAL    = {IEEE RA-L})

@article{2009_robot_learning_survey,
  author       = {Brenna D. Argall and
                  Sonia Chernova and
                  Manuela M. Veloso and
                  Brett Browning},
  title        = {A survey of robot learning from demonstration},
  journal      = {Robotics Auton. Syst.},
  year         = {2009},
}

@article{2013_robot_learning_rl_survey,
  author       = {Jens Kober and
                  J. Andrew Bagnell and
                  Jan Peters},
  title        = {Reinforcement learning in robotics: {A} survey},
  journal      = {IJRR},
  year         = {2013},
}

@article{2026_robot_learning_vla_survey,
  author       = {Yueen Ma and
                  Zixing Song and
                  Yuzheng Zhuang and
                  Jianye Hao and
                  Irwin King},
  title        = {A Survey on Vision-Language-Action Models for Embodied {AI}},
  journal      = {TNNLS},
  year         = {2026},
}

@article{2026_robot_learning_agentic_survey,
  title={Towards agentic robots: Model, data, environment, and harness},
  author={Qin, Chonghan and Wu, Dongming and Feng, Kaituo and Wang, Jiaxu and Yu, Xinlei and Sun, Peiwen and Fan, Yijia and He, Xilin and Chen, Bingao and Lian, Weitong and others},
  journal={arXiv preprint arXiv:2610.20442},
  year={2026}
}

@article{2023_qwen,
  title={Qwen technical report},
  author={Bai, Jinze and Bai, Shuai and Chu, Yunfei and Cui, Zeyu and Dang, Kai and Deng, Xiaodong and Fan, Yang and Ge, Wenbin and Han, Yu and Huang, Fei and others},
  journal={arXiv preprint arXiv:2309.16609},
  year={2023}
}

@article{2024_llama3,
  title={The llama 3 herd of models},
  author={Grattafiori, Aaron and Dubey, Abhimanyu and Jauhri, Abhinav and Pandey, Abhinav and Kadian, Abhishek and Al-Dahle, Ahmad and Letman, Aiesha and Mathur, Akhil and Schelten, Alan and Vaughan, Alex and others},
  journal={arXiv preprint arXiv:2407.21783},
  year={2024}
}

@article{1986_layered_control,
  author       = {Rodney A. Brooks},
  title        = {A robust layered control system for a mobile robot},
  journal      = {{IEEE} J. Robotics Autom.},
  year         = {1986},
}

@article{2012_OMPL,
  author       = {Ioan Alexandru Sucan and
                  Mark Moll and
                  Lydia E. Kavraki},
  title        = {The Open Motion Planning Library},
  journal      = {{IEEE} Robotics Autom. Mag.},
  year         = {2012},
}

@article{2023_AnyGrasp,
  author       = {Haoshu Fang and
                  Chenxi Wang and
                  Hongjie Fang and
                  Minghao Gou and
                  Jirong Liu and
                  Hengxu Yan and
                  Wenhai Liu and
                  Yichen Xie and
                  Cewu Lu},
  title        = {AnyGrasp: Robust and Efficient Grasp Perception in Spatial and Temporal
                  Domains},
  journal      = {TRO},
  year         = {2023},
}

@inproceedings{2024_EconomicGrasp,
  author       = {Xiao{-}Ming Wu and
                  Jia{-}Feng Cai and
                  Jian{-}Jian Jiang and
                  Dian Zheng and
                  Yi{-}Lin Wei and
                  Wei{-}Shi Zheng},
  editor       = {Ales Leonardis and
                  Elisa Ricci and
                  Stefan Roth and
                  Olga Russakovsky and
                  Torsten Sattler and
                  G{\"{u}}l Varol},
  title        = {An Economic Framework for 6-DoF Grasp Detection},
  booktitle    = {ECCV},
  year         = {2024},
}

@article{2025_motiongrasp,
  author       = {Nuo Chen and
                  Xiao{-}Ming Wu and
                  Guohao Xu and
                  Jian{-}Jian Jiang and
                  Zibo Chen and
                  Wei{-}Shi Zheng},
  title        = {MotionGrasp: Long-Term Grasp Motion Tracking for Dynamic Grasping},
  journal      = {RAL},
  year         = {2025},
}

@inproceedings{2023_SAM,
  author       = {Alexander Kirillov and
                  Eric Mintun and
                  Nikhila Ravi and
                  Hanzi Mao and
                  Chlo{\'{e}} Rolland and
                  Laura Gustafson and
                  Tete Xiao and
                  Spencer Whitehead and
                  Alexander C. Berg and
                  Wan{-}Yen Lo and
                  Piotr Doll{\'{a}}r and
                  Ross B. Girshick},
  title        = {Segment Anything},
  booktitle    = {ICCV},
  year         = {2023},
}

@inproceedings{2025_SAM2,
  author       = {Nikhila Ravi and
                  Valentin Gabeur and
                  Yuan{-}Ting Hu and
                  Ronghang Hu and
                  Chaitanya Ryali and
                  Tengyu Ma and
                  Haitham Khedr and
                  Roman R{\"{a}}dle and
                  Chlo{\'{e}} Rolland and
                  Laura Gustafson and
                  Eric Mintun and
                  Junting Pan and
                  Kalyan Vasudev Alwala and
                  Nicolas Carion and
                  Chao{-}Yuan Wu and
                  Ross B. Girshick and
                  Piotr Doll{\'{a}}r and
                  Christoph Feichtenhofer},
  title        = {{SAM} 2: Segment Anything in Images and Videos},
  booktitle    = {ICLR},
  year         = {2025},
}

@inproceedings{2026_SAM3,
  author       = {Nicolas Carion and
                  Laura Gustafson and
                  Yuan{-}Ting Hu and
                  Shoubhik Debnath and
                  Ronghang Hu and
                  Didac Suris Coll{-}Vinent and
                  Chaitanya Ryali and
                  Kalyan Vasudev Alwala and
                  Haitham Khedr and
                  Andrew Huang and
                  Jie Lei and
                  Tengyu Ma and
                  Baishan Guo and
                  Arpit Kalla and
                  Markus Marks and
                  Joseph Greer and
                  Meng Wang and
                  Peize Sun and
                  Roman R{\"{a}}dle and
                  Triantafyllos Afouras and
                  Effrosyni Mavroudi and
                  Katherine Xu and
                  Tsung{-}Han Wu and
                  Yu Zhou and
                  Liliane Momeni and
                  Rishi Hazra and
                  Shuangrui Ding and
                  Sagar Vaze and
                  Francois Porcher and
                  Feng Li and
                  Siyuan Li and
                  Aishwarya Kamath and
                  Ho Kei Cheng and
                  Piotr Doll{\'{a}}r and
                  Nikhila Ravi and
                  Kate Saenko and
                  Pengchuan Zhang and
                  Christoph Feichtenhofer},
  editor       = {Carl Vondrick and
                  Bharath Hariharan and
                  Colin Raffel and
                  Lerrel Pinto and
                  Diyi Yang and
                  Aleksandra Faust},
  title        = {{SAM} 3: Segment Anything with Concepts},
  booktitle    = {ICLR},
  year         = {2026},
}

@inproceedings{2022_owl-vit,
  title={Simple open-vocabulary object detection},
  author={Minderer, Matthias and Gritsenko, Alexey and Stone, Austin and Neumann, Maxim and Weissenborn, Dirk and Dosovitskiy, Alexey and Mahendran, Aravindh and Arnab, Anurag and Dehghani, Mostafa and Shen, Zhuoran and others},
  booktitle={ECCV},
  year={2022},
}

@inproceedings{2023_CuRobo,
  author       = {Balakumar Sundaralingam and
                  Siva Kumar Sastry Hari and
                  Adam Fishman and
                  Caelan Reed Garrett and
                  Karl Van Wyk and
                  Valts Blukis and
                  Alexander Millane and
                  Helen Oleynikova and
                  Ankur Handa and
                  Fabio Ramos and
                  Nathan D. Ratliff and
                  Dieter Fox},
  title        = {CuRobo: Parallelized Collision-Free Robot Motion Generation},
  booktitle    = {ICRA},
  year         = {2023},
}

@article{2014_moveit,
  title={Reducing the barrier to entry of complex robotic software: a moveit! case study},
  author={Coleman, David and Sucan, Ioan and Chitta, Sachin and Correll, Nikolaus},
  journal={arXiv preprint arXiv:1404.3785},
  year={2014}
}

@inproceedings{2022_PerAct,
  author       = {Mohit Shridhar and
                  Lucas Manuelli and
                  Dieter Fox},
  title        = {Perceiver-Actor: {A} Multi-Task Transformer for Robotic Manipulation},
  booktitle    = {CoRL},
  year         = {2022},
}

@inproceedings{2023_RT-1,
  author       = {Anthony Brohan and
                  Noah Brown and
                  Justice Carbajal and
                  Yevgen Chebotar and
                  Joseph Dabis and
                  Chelsea Finn and
                  Keerthana Gopalakrishnan and
                  Karol Hausman and
                  Alexander Herzog and
                  Jasmine Hsu and
                  Julian Ibarz and
                  Brian Ichter and
                  Alex Irpan and
                  Tomas Jackson and
                  Sally Jesmonth and
                  Nikhil J. Joshi and
                  Ryan Julian and
                  Dmitry Kalashnikov and
                  Yuheng Kuang and
                  Isabel Leal and
                  Kuang{-}Huei Lee and
                  Sergey Levine and
                  Yao Lu and
                  Utsav Malla and
                  Deeksha Manjunath and
                  Igor Mordatch and
                  Ofir Nachum and
                  Carolina Parada and
                  Jodilyn Peralta and
                  Emily Perez and
                  Karl Pertsch and
                  Jornell Quiambao and
                  Kanishka Rao and
                  Michael S. Ryoo and
                  Grecia Salazar and
                  Pannag R. Sanketi and
                  Kevin Sayed and
                  Jaspiar Singh and
                  Sumedh Sontakke and
                  Austin Stone and
                  Clayton Tan and
                  Huong T. Tran and
                  Vincent Vanhoucke and
                  Steve Vega and
                  Quan Vuong and
                  Fei Xia and
                  Ted Xiao and
                  Peng Xu and
                  Sichun Xu and
                  Tianhe Yu and
                  Brianna Zitkovich},
  title        = {{RT-1:} Robotics Transformer for Real-World Control at Scale},
  booktitle    = {RSS},
  year         = {2023},
}

@article{2025_diffusion-policy,
  author       = {Cheng Chi and
                  Zhenjia Xu and
                  Siyuan Feng and
                  Eric Cousineau and
                  Yilun Du and
                  Benjamin Burchfiel and
                  Russ Tedrake and
                  Shuran Song},
  title        = {Diffusion policy: Visuomotor policy learning via action diffusion},
  journal      = {IJRR},
  year         = {2025},
}

@inproceedings{2023_ACT,
  author       = {Tony Z. Zhao and
                  Vikash Kumar and
                  Sergey Levine and
                  Chelsea Finn},
  title        = {Learning Fine-Grained Bimanual Manipulation with Low-Cost Hardware},
  booktitle    = {RSS},
  year         = {2023},
}

@inproceedings{2024_Octo,
  author       = {Dibya Ghosh and
                  Homer Rich Walke and
                  Karl Pertsch and
                  Kevin Black and
                  Oier Mees and
                  Sudeep Dasari and
                  Joey Hejna and
                  Tobias Kreiman and
                  Charles Xu and
                  Jianlan Luo and
                  You Liang Tan and
                  Lawrence Yunliang Chen and
                  Quan Vuong and
                  Ted Xiao and
                  Pannag R. Sanketi and
                  Dorsa Sadigh and
                  Chelsea Finn and
                  Sergey Levine},
  title        = {Octo: An Open-Source Generalist Robot Policy},
  booktitle    = {RSS},
  year         = {2024},
}

@inproceedings{2023_RT-2,
  author       = {Brianna Zitkovich and
                  Tianhe Yu and
                  Sichun Xu and
                  Peng Xu and
                  Ted Xiao and
                  Fei Xia and
                  Jialin Wu and
                  Paul Wohlhart and
                  Stefan Welker and
                  Ayzaan Wahid and
                  Quan Vuong and
                  Vincent Vanhoucke and
                  Huong T. Tran and
                  Radu Soricut and
                  Anikait Singh and
                  Jaspiar Singh and
                  Pierre Sermanet and
                  Pannag R. Sanketi and
                  Grecia Salazar and
                  Michael S. Ryoo and
                  Krista Reymann and
                  Kanishka Rao and
                  Karl Pertsch and
                  Igor Mordatch and
                  Henryk Michalewski and
                  Yao Lu and
                  Sergey Levine and
                  Lisa Lee and
                  Tsang{-}Wei Edward Lee and
                  Isabel Leal and
                  Yuheng Kuang and
                  Dmitry Kalashnikov and
                  Ryan Julian and
                  Nikhil J. Joshi and
                  Alex Irpan and
                  Brian Ichter and
                  Jasmine Hsu and
                  Alexander Herzog and
                  Karol Hausman and
                  Keerthana Gopalakrishnan and
                  Chuyuan Fu and
                  Pete Florence and
                  Chelsea Finn and
                  Kumar Avinava Dubey and
                  Danny Driess and
                  Tianli Ding and
                  Krzysztof Marcin Choromanski and
                  Xi Chen and
                  Yevgen Chebotar and
                  Justice Carbajal and
                  Noah Brown and
                  Anthony Brohan and
                  Montserrat Gonzalez Arenas and
                  Kehang Han},
  editor       = {Jie Tan and
                  Marc Toussaint and
                  Kourosh Darvish},
  title        = {{RT-2:} Vision-Language-Action Models Transfer Web Knowledge to Robotic
                  Control},
  booktitle    = {CoRL},
  year         = {2023},
}

@inproceedings{2024_OpenVLA,
  author       = {Moo Jin Kim and
                  Karl Pertsch and
                  Siddharth Karamcheti and
                  Ted Xiao and
                  Ashwin Balakrishna and
                  Suraj Nair and
                  Rafael Rafailov and
                  Ethan Paul Foster and
                  Pannag R. Sanketi and
                  Quan Vuong and
                  Thomas Kollar and
                  Benjamin Burchfiel and
                  Russ Tedrake and
                  Dorsa Sadigh and
                  Sergey Levine and
                  Percy Liang and
                  Chelsea Finn},
  title        = {OpenVLA: An Open-Source Vision-Language-Action Model},
  booktitle    = {CoRL},
  year         = {2024},
}

@article{2024_paligemma,
  title={Paligemma: A versatile 3b vlm for transfer},
  author={Beyer, Lucas and Steiner, Andreas and Pinto, Andr{\'e} Susano and Kolesnikov, Alexander and Wang, Xiao and Salz, Daniel and Neumann, Maxim and Alabdulmohsin, Ibrahim and Tschannen, Michael and Bugliarello, Emanuele and others},
  journal={arXiv preprint arXiv:2407.07726},
  year={2024}
}

@article{2024_pi0,
  author       = {Kevin Black and
                  Noah Brown and
                  Danny Driess and
                  Adnan Esmail and
                  Michael Equi and
                  Chelsea Finn and
                  Niccolo Fusai and
                  Lachy Groom and
                  Karol Hausman and
                  Brian Ichter and
                  Szymon Jakubczak and
                  Tim Jones and
                  Liyiming Ke and
                  Sergey Levine and
                  Adrian Li{-}Bell and
                  Mohith Mothukuri and
                  Suraj Nair and
                  Karl Pertsch and
                  Lucy Xiaoyang Shi and
                  James Tanner and
                  Quan Vuong and
                  Anna Walling and
                  Haohuan Wang and
                  Ury Zhilinsky},
  title        = {{\(\pi\)}\({}_{\mbox{0}}\): {A} Vision-Language-Action Flow Model
                  for General Robot Control},
  journal      = {arXiv preprint arXiv:2410.24164},
  year         = {2024},
}

@article{2025_fast,
  title={Fast: Efficient action tokenization for vision-language-action models},
  author={Pertsch, Karl and Stachowicz, Kyle and Ichter, Brian and Driess, Danny and Nair, Suraj and Vuong, Quan and Mees, Oier and Finn, Chelsea and Levine, Sergey},
  journal={arXiv preprint arXiv:2501.09747},
  year={2025}
}

@article{2025_pi05,
  title={{$\pi_{0.5}$}: A Vision-Language-Action Model with Open-World Generalization},
  author={Intelligence, Physical and Black, Kevin and Brown, Noah and Darpinian, James and Dhabalia, Karan and Driess, Danny and Esmail, Adnan and Equi, Michael and Finn, Chelsea and Fusai, Niccolo and others},
  journal={arXiv preprint arXiv:2504.16054},
  year={2025}
}

@article{2025_gr00t,
  title={Gr00t n1: An open foundation model for generalist humanoid robots},
  author={Bjorck, Johan and Casta{\~n}eda, Fernando and Cherniadev, Nikita and Da, Xingye and Ding, Runyu and Fan, Linxi and Fang, Yu and Fox, Dieter and Hu, Fengyuan and Huang, Spencer and others},
  journal={arXiv preprint arXiv:2503.14734},
  year={2025}
}

@inproceedings{2026_VLANeXt,
  author       = {Xiao{-}Ming Wu and
                  Bin Fan and
                  Kang Liao and
                  Jian{-}Jian Jiang and
                  Runze Yang and
                  Yihang Luo and
                  Zhonghua Wu and
                  Wei{-}Shi Zheng and
                  Chen Change Loy},
  title        = {VLANeXt: Recipes for Building Strong {VLA} Models},
  booktitle      = {ICML},
  year         = {2026},
}

@article{2026_cosmos3,
  title={Cosmos 3: Omnimodal world models for physical ai},
  author={Agarwal, Niket and Ali, Arslan and Allen, Jon and Antolini, Martin and Aubame, Adeline and Azzolini, Alisson and Bai, Junjie and Bala, Maciej and Balaji, Yogesh and Bapst, Josh and others},
  journal={arXiv preprint arXiv:2606.02800},
  year={2026}
}

@misc{2026_flux-3-action,
  author = {Forest Labs, Black},
  title = {FLUX 3 Action},
  url = {https://bfl.ai/models/flux-3-action},
  year = {2026}
}

@article{2026_atomic-wam,
  title={Atomic-WAM: Execution-Grounded Atomic-Motion Reasoning for World Action Models},
  author={Shi, Shijun and Liu, Zhe and Hou, Zhi and Chen, Yinan and Zhang, Tao and Hu, Kai and Liu, Yong and Zhang, Jiangning},
  journal={arXiv preprint arXiv:2610.20157},
  year={2026}
}

@article{2026_lingbot-va,
  title={Causal world modeling for robot control},
  author={Li, Lin and Zhang, Qihang and Luo, Yiming and Yang, Shuai and Wang, Ruilin and Han, Fei and Yu, Mingrui and Gao, Zelin and Xue, Nan and Zhu, Xing and others},
  journal={arXiv preprint arXiv:2601.21998},
  year={2026}
}

@article{2026_dreamzero,
  title={World action models are zero-shot policies},
  author={Ye, Seonghyeon and Ge, Yunhao and Zheng, Kaiyuan and Gao, Shenyuan and Yu, Sihyun and Kurian, George and Indupuru, Suneel and Tan, You Liang and Zhu, Chuning and Xiang, Jiannan and others},
  journal={arXiv preprint arXiv:2602.15922},
  year={2026}
}

@article{2026_fast-wam,
  title={Fast-wam: Do world action models need test-time future imagination?},
  author={Yuan, Tianyuan and Dong, Zibin and Liu, Yicheng and Zhao, Hang},
  journal={arXiv preprint arXiv:2603.16666},
  year={2026}
}

@inproceedings{2022_zero-shot-planner,
  author       = {Wenlong Huang and
                  Pieter Abbeel and
                  Deepak Pathak and
                  Igor Mordatch},
  title        = {Language Models as Zero-Shot Planners: Extracting Actionable Knowledge
                  for Embodied Agents},
  booktitle    = {ICML},
  year         = {2022},
}

@inproceedings{2023_progprompt,
  author       = {Ishika Singh and
                  Valts Blukis and
                  Arsalan Mousavian and
                  Ankit Goyal and
                  Danfei Xu and
                  Jonathan Tremblay and
                  Dieter Fox and
                  Jesse Thomason and
                  Animesh Garg},
  title        = {ProgPrompt: Generating Situated Robot Task Plans using Large Language
                  Models},
  booktitle    = {ICRA},
  year         = {2023},
}

@article{2023_instruct2act,
  title={Instruct2act: Mapping multi-modality instructions to robotic actions with large language model},
  author={Huang, Siyuan and Jiang, Zhengkai and Dong, Hao and Qiao, Yu and Gao, Peng and Li, Hongsheng},
  journal={arXiv preprint arXiv:2305.11176},
  year={2023}
}

@inproceedings{2023_code-as-policies,
  author       = {Jacky Liang and
                  Wenlong Huang and
                  Fei Xia and
                  Peng Xu and
                  Karol Hausman and
                  Brian Ichter and
                  Pete Florence and
                  Andy Zeng},
  title        = {Code as Policies: Language Model Programs for Embodied Control},
  booktitle    = {ICRA},
  year         = {2023},
}

@inproceedings{2023_voxposer,
  author       = {Wenlong Huang and
                  Chen Wang and
                  Ruohan Zhang and
                  Yunzhu Li and
                  Jiajun Wu and
                  Li Fei{-}Fei},
  title        = {VoxPoser: Composable 3D Value Maps for Robotic Manipulation with Language
                  Models},
  booktitle    = {CoRL},
  year         = {2023},
}

@inproceedings{2024_Statler,
  author       = {Takuma Yoneda and
                  Jiading Fang and
                  Peng Li and
                  Huanyu Zhang and
                  Tianchong Jiang and
                  Shengjie Lin and
                  Ben Picker and
                  David Yunis and
                  Hongyuan Mei and
                  Matthew R. Walter},
  title        = {Statler: State-Maintaining Language Models for Embodied Reasoning},
  booktitle    = {ICRA},
  year         = {2024},
}

@inproceedings{2024_ReKep,
  author       = {Wenlong Huang and
                  Chen Wang and
                  Yunzhu Li and
                  Ruohan Zhang and
                  Li Fei{-}Fei},
  title        = {ReKep: Spatio-Temporal Reasoning of Relational Keypoint Constraints
                  for Robotic Manipulation},
  booktitle    = {CoRL},
  year         = {2024},
}

@article{2024_llm-zero-shot-traj,
  author       = {Teyun Kwon and
                  Norman Di Palo and
                  Edward Johns},
  title        = {Language Models as Zero-Shot Trajectory Generators},
  journal      = {RA-L},
  year         = {2024},
}

@inproceedings{2026_dream2flow,
  author       = {Karthik Dharmarajan and
                  Wenlong Huang and
                  Jiajun Wu and
                  Li Fei{-}Fei and
                  Ruohan Zhang},
  title        = {Dream2Flow: Bridging Video Generation and Open-World Manipulation
                  with 3D Object Flow},
  booktitle    = {ICRA},
  year         = {2026},
}

@inproceedings{2026_cap-x,
  title={CaP-X: A framework for benchmarking and improving coding agents for robot manipulation},
  author={Fu, Letian and Yu, Justin and El-Refai, Karim and Kou, Ethan and Xue, Haoru and Huang, Huang and Xiao, Wenli and Wang, Guanzhi and Niu, Dantong and Li, Fei-Fei and others},
  booktitle={ICML},
  year={2026}
}

@article{2026_openeta,
  title={ETA: A new agentic paradigm for embodied tasks},
  author={Chen, Yitong and Huai, Zezheng and Li, Sixian and Wang, Yubang and Zhang, Haozhe and Zhang, Yifei and Chen, Hechang and Gong, Jingjing and Jiang, Yu-Gang and Qiu, Xipeng},
  journal={arXiv preprint arXiv:2608.03924},
  year={2026}
}

@article{2026_RPG,
  title={Reconstruct, practice, go real: Guided self-improvement for embodied agents},
  author={Wang, Yen-Jen and Jiang, Haozhe and Deng, Shuying and Xue, Haoru and Ye, Weirui and Duan, Rocky and Haghtalab, Nika and Sastry, S Shankar and Abbeel, Pieter and Qi, Haozhi},
  journal={arXiv preprint arXiv:2610.02204},
  year={2026}
}

@article{2026_embodied-turing-machines,
  title={Embodied Turing Machines: Stateful Code for Robot Recursive Self-Improvement},
  author={Hu, Kairui and Hu, Siyuan and Hong, Fangzhou and Chen, Zhaoxi and Liu, Ziwei},
  journal={arXiv preprint arXiv:2610.12369},
  year={2026}
}

@article{2026_thinklocomanip,
  title={ThinkLocoManip: A Whole-Body Control Agentic Harness for Legged Loco-Manipulation},
  author={Chen, Nuo and Zhang, Shizeng and Yang, Yujie and Wang, Jinghong and Qu, Zhenyang and Wu, Xiao-Ming and Guo, Yunfei and Lu, Zhongwen and Liu, Haojie and Lu, Chunpeng and others},
  journal={arXiv preprint arXiv:2610.34370},
  year={2026}
}

@article{2026_mcga,
  title={Make code as policy great again: Frontier agents write, call, and evolve robot tools},
  author={Ge, Shijia and Zhou, Alex and Zeng, Jianshu and Wan, Yexing and Wu, Di and Zheng, Zelin and Wang, Yazhe and Jia, Zhiqi and Shangguan, Xuan and Zhu, Jay and others},
  journal={arXiv preprint arXiv:2609.39018},
  year={2026}
}

@article{2026_captain,
  title={CaPtain: A Pipeline for Post-Training Code-as-Policy Robot Agents},
  author={Fan, Yijia and He, Xilin and Jiang, Yilei and Wu, Dongming and Wang, Jiaxu and Yue, Xiangyu},
  journal={arXiv preprint arXiv:2610.40799},
  year={2026}
}

@article{2026_AMP1,
  title={On Agentic Manipulation Policies},
  author={Chen, Chen and Bagherieh, Omid and Balaji, Naveen Nikamanth Appiah and Gonzalez, Jesse T and Heravi, Negin and Islam, Fahad and Li, Xueting and Regli, Jean-Baptiste and Ren, Kejia and {\v{S}}vec, Petr and others},
  journal={arXiv preprint arXiv:2611.09892},
  year={2026}
}

@inproceedings{2022_saycan,
  author       = {Brian Ichter and
                  Anthony Brohan and
                  Yevgen Chebotar and
                  Chelsea Finn and
                  Karol Hausman and
                  Alexander Herzog and
                  Daniel Ho and
                  Julian Ibarz and
                  Alex Irpan and
                  Eric Jang and
                  Ryan Julian and
                  Dmitry Kalashnikov and
                  Sergey Levine and
                  Yao Lu and
                  Carolina Parada and
                  Kanishka Rao and
                  Pierre Sermanet and
                  Alexander Toshev and
                  Vincent Vanhoucke and
                  Fei Xia and
                  Ted Xiao and
                  Peng Xu and
                  Mengyuan Yan and
                  Noah Brown and
                  Michael Ahn and
                  Omar Cortes and
                  Nicolas Sievers and
                  Clayton Tan and
                  Sichun Xu and
                  Diego Reyes and
                  Jarek Rettinghouse and
                  Jornell Quiambao and
                  Peter Pastor and
                  Linda Luu and
                  Kuang{-}Huei Lee and
                  Yuheng Kuang and
                  Sally Jesmonth and
                  Nikhil J. Joshi and
                  Kyle Jeffrey and
                  Rosario Jauregui Ruano and
                  Jasmine Hsu and
                  Keerthana Gopalakrishnan and
                  Byron David and
                  Andy Zeng and
                  Chuyuan Kelly Fu},
  editor       = {Karen Liu and
                  Dana Kulic and
                  Jeffrey Ichnowski},
  title        = {Do As {I} Can, Not As {I} Say: Grounding Language in Robotic Affordances},
  booktitle    = {CoRL},
  year         = {2022},
}

@inproceedings{2022_Inner-Monologue,
  author       = {Wenlong Huang and
                  Fei Xia and
                  Ted Xiao and
                  Harris Chan and
                  Jacky Liang and
                  Pete Florence and
                  Andy Zeng and
                  Jonathan Tompson and
                  Igor Mordatch and
                  Yevgen Chebotar and
                  Pierre Sermanet and
                  Tomas Jackson and
                  Noah Brown and
                  Linda Luu and
                  Sergey Levine and
                  Karol Hausman and
                  Brian Ichter},
  editor       = {Karen Liu and
                  Dana Kulic and
                  Jeffrey Ichnowski},
  title        = {Inner Monologue: Embodied Reasoning through Planning with Language
                  Models},
  booktitle    = {CoRL},
  year         = {2022},
}

@inproceedings{2023_MOO,
  author       = {Austin Stone and
                  Ted Xiao and
                  Yao Lu and
                  Keerthana Gopalakrishnan and
                  Kuang{-}Huei Lee and
                  Quan Vuong and
                  Paul Wohlhart and
                  Sean Kirmani and
                  Brianna Zitkovich and
                  Fei Xia and
                  Chelsea Finn and
                  Karol Hausman},
  title        = {Open-World Object Manipulation using Pre-Trained Vision-Language Models},
  booktitle    = {CoRL},
  year         = {2023},
}

@inproceedings{2025_Hi-Robot,
  author       = {Lucy Xiaoyang Shi and
                  Brian Ichter and
                  Michael Robert Equi and
                  Liyiming Ke and
                  Karl Pertsch and
                  Quan Vuong and
                  James Tanner and
                  Anna Walling and
                  Haohuan Wang and
                  Niccolo Fusai and
                  Adrian Li{-}Bell and
                  Danny Driess and
                  Lachy Groom and
                  Sergey Levine and
                  Chelsea Finn},
  editor       = {Aarti Singh and
                  Maryam Fazel and
                  Daniel Hsu and
                  Simon Lacoste{-}Julien and
                  Felix Berkenkamp and
                  Tegan Maharaj and
                  Kiri Wagstaff and
                  Jerry Zhu},
  title        = {Hi Robot: Open-Ended Instruction Following with Hierarchical Vision-Language-Action
                  Models},
  booktitle    = {ICML},
  publisher    = {{PMLR} / OpenReview.net},
  year         = {2025},
}

@inproceedings{2024_doremi,
  author       = {Yanjiang Guo and
                  Yen{-}Jen Wang and
                  Lihan Zha and
                  Jianyu Chen},
  title        = {DoReMi: Grounding Language Model by Detecting and Recovering from
                  Plan-Execution Misalignment},
  booktitle    = {IROS},
  year         = {2024},
}

@article{2026_harness-vla,
  title={Harness vla: Steering frozen vlas into reliable manipulation primitives via memory-guided agents},
  author={Zhang, Yixian and Zhang, Huanming and Gao, Feng and Li, Xiao and Liu, Zhihao and Nie, Yi and Zhu, Chunyang and Qiu, Jiaxing and Yan, Yuchen and Liu, Jiyuan and others},
  journal={arXiv preprint arXiv:2607.08448},
  year={2026}
}

@article{2026_volo,
  title={VoLo: A Physical Orchestrator for Open-Vocabulary Long-Horizon Manipulation},
  author={Chen, Siyi and Hadfield, Hugo and Zook, Alex and Uy, Mikaela Angelina and Song, Chan Hee and Coumans, Erwin and Yang, Xuning and Ladhak, Faisal and Qu, Qing and Birchfield, Stan and others},
  journal={arXiv preprint arXiv:2606.07723},
  year={2026}
}

@article{2026_PyRUA-Learn,
  title={Fewer tokens, better action: {GPT}-6 Astra robot agents with 14\% higher success rate but 65\% fewer tokens},
  author={Si, Ruiyang and Bi, Jianxin and Yang, Shunyu and Ni, Rui and Huang, Wenbo and Wang, Qiang and Jiang, Shulong and Wang, Duomin and Li, Xiuyu and Feng, Haiwen and others},
  journal={arXiv preprint arXiv:2610.01939},
  year={2026}
}

@misc{2026_gpt6,
  title  = {{{GPT}-6 and Intelligent UI for Everyone}},
  author = {{OpenAI}},
  year   = {2026},
  url    = {https://openai.com/index/gpt-6-for-everyone/}
}

@article{2026_show-harness,
  title={Show-Harness: Just a VLM agent can play robots},
  author={Chen, Yanzhe and Bai, Zechen and Cao, Zhijun and Zeng, Wenzheng and Lin, Kevin Qinghong and Lin, Yiqi and Liang, Guoqiang and Ma, Kevin Yuchen and Huang, Qiming and Shou, Mike Zheng},
  journal={arXiv preprint arXiv:2609.10522},
  year={2026}
}

@article{2026_gpt-6-astra-robodojo,
  title={An Unexpected Robot Policy: Early Evaluations of {GPT}-6 {Astra} on {RoboDojo} and Beyond},
  author={Zhang, Wenbo and Wang, Kaixuan and Ouyang, Yutao and Huang, Xiaoyu and Li, Liyang and Su, Kailun and Jin, Weiyang and Chai, Wenhao and Liang, Haotian and Dou, Zhiyang and others},
  journal={arXiv preprint arXiv:2609.24170},
  year={2026}
}

@article{2026_gpt-as-policy,
  title={Systematically exploring the capabilities of {GPT}-6 {Astra} as embodied policies},
  author={Team, Galbot and Chen, Xuchuan and Cheng, Xiaoqian and Deng, Yu and Ding, Lihe and Dong, Shaocong and Gao, Xiangjun and Jia, Haozhe and Li, Zekai and Li, Zhoujian and others},
  journal={arXiv preprint arXiv:2609.38537},
  year={2026}
}

@article{2026_GPT-Policy,
  title={In-context robot learning with VLM agents},
  author={Cheng, Dongzhou and Yi, Taoran and Fang, Ye and Zhang, Xingwu and Feng, Fan and Li, Yixuan and Zhuang, Gengxiong and Wang, Rongze and Yang, Shuai and Song, Wei and others},
  journal={arXiv preprint arXiv:2609.19138},
  year={2026}
}

@article{2026_spotter,
  title={Spotter: Let the Embodied Model Lead, and the VLM Reflect for It},
  author={Li, Long and Zhao, Qichao and Yang, Yue and Xu, Fan and Wang, Zhe and Liew, Alan Wee-Chung and Qu, Chao and Shen, Heng Tao and Pan, Shirui},
  journal={arXiv preprint arXiv:2609.36808},
  year={2026}
}

@article{2026_physevo,
  title={{PhysEvo}: {Astra} Can Act, Let It},
  author={Tian, Wenqing and Zhang, Zeyu and Liu, Zhaocheng and Liu, Fengwei and Liu, Qiang and Wang, Liang},
  journal={arXiv preprint arXiv:2610.08995},
  year={2026}
}

@article{2026_roborsi,
  title={Has {GPT}-6-{Astra} Achieved Embodied Intelligence?},
  author={Team, Joint Research and Cai, Zhongang and Cao, Ziang and Chen, Tailai and Chua, Han Chong and Dai, Linjun and Deng, Hanming and Deng, Hokin and Diao, Haiwen and Ding, Chenjing and others},
  journal={arXiv preprint arXiv:2610.22471},
  year={2026}
}

@article{2026_bridging-frontier,
  title={Bridging Frontier Reasoning and Robot Execution: From Autonomous Demonstration Generation to Dense Language Supervision},
  author={Kim, Bosung and Trevithick, Alexander and Wang, Ruiyi and Ammanabrolu, Prithviraj},
  journal={arXiv preprint arXiv:2610.03615},
  year={2026}
}

@article{2026_fastharness,
  title={FastHarness: Experience-Driven Review Scheduling for Vision-Language-Action Policies},
  author={Lin, Tao and Du, Yuxin and Wang, Haoxiao and Zhong, Zile and Zhao, Bo},
  journal={arXiv preprint arXiv:2610.31427},
  year={2026}
}

@article{2026_astra-ebench,
  title={Profiling a frontier agent with EBench: From direct control to harness with policy},
  author={Yao, Yifei and Zou, Yude and Jiao, Guanbo and Gao, Xing and Gao, Ning and Wang, Yukai and Chen, Jiantong and Zhu, Zihou and Li, Xinyu and Yan, Ningyu and others},
  journal={arXiv preprint arXiv:2610.35832},
  year={2026}
}

@article{2026_spatialharness,
  title={SpatialHarness: Test-Time Spatial Scaffolding for Fine Robotic Manipulation},
  author={Wang, Jiayu and Yu, Yue and Zhu, Bin and Yang, Zhiyao and Chen, Jingjing},
  journal={arXiv preprint arXiv:2610.12457},
  year={2026}
}

@misc{2026_astranomy,
  title        = {Astranomy: The study of {GPT-6 Astra} as a native robot policy},
  author       = {Yang, Jenai Xuning and Khetan, Ashna and Zhang, Shun},
  year         = {2026},
  howpublished = {\url{https://research.nvidia.com/labs/srl/projects/astranomy/}},
  note         = {Blog post},
}

@inproceedings{2023_LIBERO,
  author       = {Bo Liu and
                  Yifeng Zhu and
                  Chongkai Gao and
                  Yihao Feng and
                  Qiang Liu and
                  Yuke Zhu and
                  Peter Stone},
  editor       = {Alice Oh and
                  Tristan Naumann and
                  Amir Globerson and
                  Kate Saenko and
                  Moritz Hardt and
                  Sergey Levine},
  title        = {{LIBERO:} Benchmarking Knowledge Transfer for Lifelong Robot Learning},
  booktitle    = {NeurIPS},
  year         = {2023},
}

@article{2025_libero-pro,
  title={Libero-pro: Towards robust and fair evaluation of vision-language-action models beyond memorization},
  author={Zhou, Xueyang and Xu, Yangming and Tie, Guiyao and Chen, Yongchao and Zhang, Guowen and Chu, Duanfeng and Zhou, Pan and Sun, Lichao},
  journal={arXiv preprint arXiv:2510.03827},
  year={2025}
}

@article{2026_robolab,
  title={Robolab: A high-fidelity simulation benchmark for analysis of task generalist policies},
  author={Yang, Jenai Xuning and Dagli, Rishit and Zook, Alex and Hadfield, Hugo and Goyal, Ankit and Birchfield, Stan and Ramos, Fabio and Tremblay, Jonathan},
  journal={arXiv preprint arXiv:2604.09860},
  year={2026}
}

@article{2026_robodojo,
  title={{RoboDojo}: A unified sim-and-real benchmark for comprehensive evaluation of generalist robot manipulation policies},
  author={Chen, Tianxing and Chen, Yue and Li, Zixuan and Tang, Junyuan and Su, Kailun and Lu, Haoran and Wan, Weijie and Chen, Baijun and Liu, Songling and Yan, Haowen and others},
  journal={arXiv preprint arXiv:2607.04434},
  year={2026}
}
}

\renewcommand{\thetitle}{Embo-Harness: Towards Understanding Frontier-Model-as-Policy Harness}
\clearpage
\setcounter{page}{1}
\maketitlesupplementary

\section{Supplementary of Embo-Harness}
\textbf{Detail of the Notice Reminder.} In our recipe exploration, we found that the notes distilled through recursive self-improvement (RSI) provide limited benefits to manipulation performance. Here, we show the note we used to illustrate its content and format. As shown below, the notes consist of general, task-agnostic reminders distilled from common failure cases, with task-specific information removed. They primarily highlight potential pitfalls in robotic manipulation that frontier models may encounter during execution. But frontier models mostly understand how to fix the error, rather than just relying on predefined notes.

\begin{tcolorbox}[
    float=h,
    colback=gray!5,
    colframe=gray!40,
    boxrule=0.4pt,
    arc=1mm,
    left=2mm,
    right=2mm,
    top=1mm,
    bottom=1mm
]
\footnotesize
\textbf{Manipulation Note:}
\begin{itemize}
    \setlength{\itemsep}{1pt}
    \setlength{\topsep}{2pt}
    \item \textbf{Plan:} Briefly plan before acting, then adjust based on feedback.
    \item \textbf{Move:} Use smooth, collision-free motions and approach targets from a stable direction. Consider larger travel distances in clear space.
    \item \textbf{Fine movement:} Use small, precise motions near objects and during contact.
    \item \textbf{Grasp \& place:} Grasp near the object's stable center; place it close to the target surface before releasing.
    \item \textbf{Push \& pull:} Apply force near a stable contact point and align the motion with the intended direction.
    \item \textbf{Track:} Track required objects, destinations, counts, and order; preserve completed work.
    \item \textbf{Observe:} Use both views to resolve uncertainty; re-localize objects after unexpected motion.
    \item \textbf{Geometry:} Match grasps to jaw span and account for hand/object offsets.
    \item \textbf{Align:} Align the object's actual pose and footprint with the required support or opening.
    \item \textbf{Verify:} Confirm pickup, release, and final stability from observations.
    \item \textbf{Recover:} Diagnose failures and change the grasp or approach when retries stall.
    \item \textbf{Budget:} Reserve enough steps for remaining transfers, releases, and withdrawal.
\end{itemize}
\end{tcolorbox}

\textbf{Detail of the Real-world Experiments.} 
We evaluate Embo-Harness on four real-world manipulation tasks: (1) block stacking, which requires stacking blocks on top of each other; (2) fruit packing, which involves placing fruits into a basket; (3) dish washing, which requires turning on the faucet of a toy sink; and (4) flower arranging, which involves inserting flowers into a vase.

\textbf{Details of the Spatial Estimation Experiments.} We evaluate spatial estimation accuracy on the same subset of 20 RoboLab tasks used in our recipe exploration. For each task, we select the target object of manipulation and ask GPT to estimate its 3D position, compared against the ground-truth position. For each setting, we run one spatial estimation trial per task. For the moving closer setting, GPT first generates and executes an action to approach the target object before estimating its 3D position.

\section{More Analysis of the Frontier Models}

\begin{table}[t]
    \centering
    \small
    \caption{\textbf{Effect of action chunk size on action displacement (20 tasks, 3 runs in RoboLab benchmark).}
    We report success rate (SR) and average action displacement
    over 60 trajectories, normalized to that of chunk size 8
    (1.00$\times$). Adaptive chunking is further analyzed
    across different selected chunk sizes.}
    \vspace{-2mm}
    \label{tab:displacement}
    \setlength{\tabcolsep}{2pt}
    \renewcommand{\arraystretch}{1.05}
    \begin{tabular*}{\columnwidth}{@{\extracolsep{\fill}}lcc@{}}
        \toprule
        \textbf{policy} & \textbf{SR (\%)} & \textbf{distance} \\
        \midrule
        \rowcolor{sectiongray}
        \multicolumn{3}{l}{\textit{fixed chunking}} \\
        chunk 4  & 40.0 & 1.01$\times$ \\
        chunk 8  & 58.3 & 1.00$\times$ \\
        chunk 15 & 43.3 & 0.95$\times$ \\
        chunk 30 & 21.7 & 1.10$\times$ \\
        \midrule
        \rowcolor{sectiongray}
        \multicolumn{3}{l}{\textit{adaptive chunking}} \\
        overall & 15.0 & 0.96$\times$ \\
        \makecell[l]{\quad chunk 1--24, 25--34\\
        \quad chunk 35--44, $\geq$45}
        & -- &
        \makecell[c]{0.56$\times$, 0.94$\times$\\
        1.32$\times$, 1.66$\times$} \\
        \bottomrule
    \end{tabular*}
    \vspace{-5mm}
\end{table}

\begin{figure}[t]
    \centering
    \includegraphics[width=\linewidth]{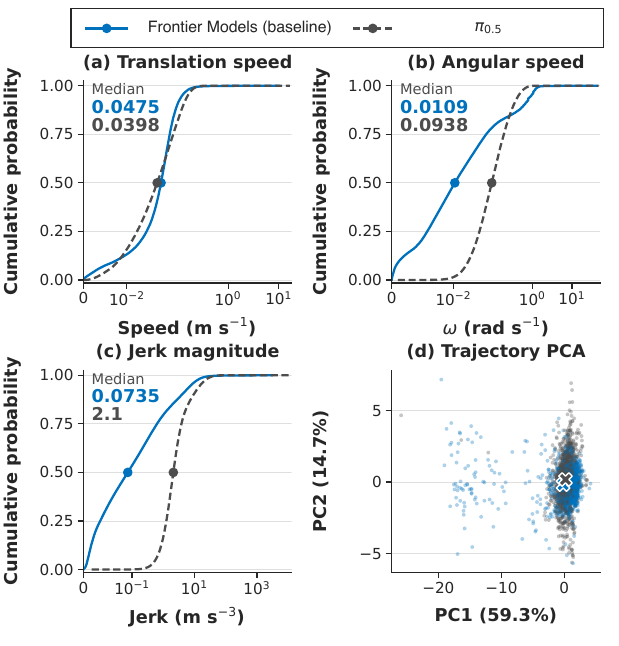}
    \vspace{-5mm}
    \caption{\textbf{Trajectory distribution difference between GPT and $\pi_{0.5}$  (120 tasks, 10 runs in RoboLab benchmark).} We compare the distributions of (a) translation speed, (b) angular speed, and (c) jerk magnitude, along with (d) PCA projections of rollout trajectories. GPT Harness exhibits a broader trajectory distribution than $\pi_{0.5}$ in the PCA projection, suggesting distinct motion patterns.}
    \vspace{-5mm}
    \label{fig:gpt_vs_pi05}
\end{figure}

\begin{figure}[t]
    \centering
    \includegraphics[width=\linewidth]{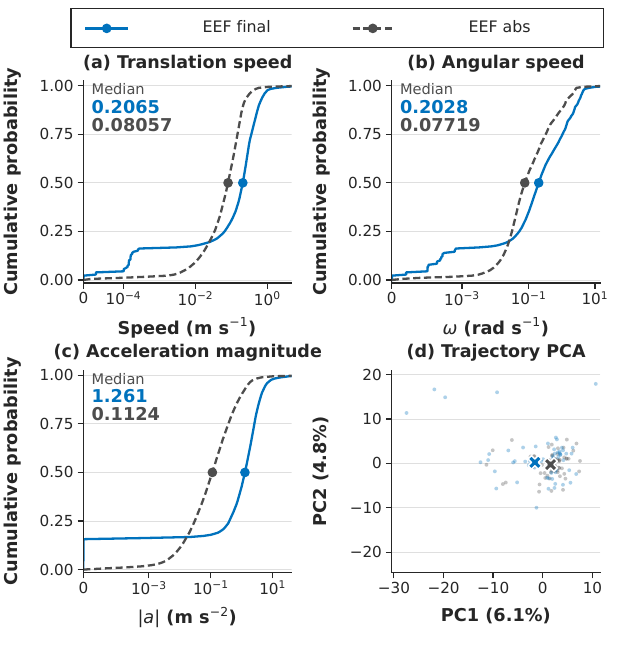}
    \vspace{-5mm}
    \caption{\textbf{Trajectory Distribution Comparison between EEF Absolute and EEF Destination Control  (20 tasks, 3 runs in RoboLab benchmark).} We compare the distributions of (a) translation speed, (b) angular speed, and (c) acceleration magnitude, along with (d) PCA projections of rollout trajectories. Despite differences in motion magnitudes caused by their execution rules, the two strategies exhibit broadly similar distributional patterns and overlapping trajectory projections, suggesting shared spatial motion characteristics.}
    \vspace{-5mm}
    \label{fig:eef_final_vs_abs}
\end{figure}

\textbf{Action Displacement Analysis.} To better understand how frontier models perceive action execution horizons, we analyze the generated action displacements under different chunking strategies. As shown in Table~\ref{tab:displacement}, despite increasing the fixed chunk size from 4 to 30, the average action displacement remains nearly unchanged, ranging from 0.95$\times$ to 1.10$\times$ relative to chunk size 8. Interestingly, adaptive chunking also exhibits a comparable overall displacement of 0.96$\times$, despite dynamically selecting different chunk sizes. These observations suggest that frontier models tend to generate similar target displacements regardless of the chunking strategy, indicating limited sensitivity to execution horizons. Overall, these findings highlight the difficulty frontier models face in consistently aligning action displacements with execution horizons, motivating the use of a short, fixed action chunk.

\textbf{Trajectory Distribution Analysis across Policy Strategies.} To further investigate the motion patterns of frontier models, we compare the rollout trajectories of GPT Harness with those of $\pi_{0.5}$~\cite{2025_pi05}, a VLA trained on low-level robotic action data. As shown in Fig.~\ref{fig:gpt_vs_pi05}, the two methods exhibit comparable translation speed distributions, with median values of 0.0475 and 0.0398 m/s, respectively. However, GPT Harness exhibits substantially lower angular speed (0.0109 vs. 0.0938 rad/s) and jerk magnitude (0.0735 vs. 2.1 m/s$^3$), although these differences may also be influenced by their action execution strategies. More importantly, the PCA visualization reveals that GPT Harness exhibits a broader trajectory distribution, whereas $\pi_{0.5}$ produces more concentrated trajectories. These observations highlight distinct motion characteristics between frontier models and conventional VLAs, providing complementary evidence that frontier models do not simply reproduce the low-level action patterns learned by imitation-based policies.

\textbf{Trajectory Distribution Analysis across Action Strategies.} To further investigate whether short-range action generation and direct destination prediction share similar spatial behaviors, we compare their rollout trajectory distributions. Specifically, EEF absolute control predicts short-range target poses, with execution velocities determined by the predicted displacements, whereas EEF destination control predicts the final target pose and moves toward it incrementally at predefined velocities. As shown in Fig.~\ref{fig:eef_final_vs_abs}, despite differences in motion magnitudes caused by their execution rules, the two strategies exhibit broadly similar distributional patterns in translation speed, angular speed, and acceleration, along with overlapping trajectory distributions in the PCA projection. These observations suggest shared spatial motion characteristics, supporting our hypothesis that frontier models' action generation is primarily guided by spatial target estimation rather than precise low-level action modeling.

\section{More Visualizations}
We further visualize representative successful rollouts of Embo-Harness from the 120 tasks in the RoboLab benchmark, and demonstrations from a diverse set of real-world manipulation tasks. All tasks are executed in a purely zero-shot setting, without task-specific training or fine-tuning. These demonstrations highlight the strong generalization and task understanding capabilities of Embo-Harness across diverse instructions and environments. The corresponding rollouts are provided in the supplementary video.

From the rollout videos in both simulation and real-world experiments, we observe a recurring behavior: frontier models first move toward the estimated target position and then attempt to grasp. Although the initial grasp often fails, subsequent corrections and adjustments frequently lead to successful execution. These observations provide further qualitative evidence that frontier models rely primarily on spatial perception and iterative correction rather than precise low-level action generation.

\begin{figure*}[t]
    \centering
    \includegraphics[width=\linewidth]{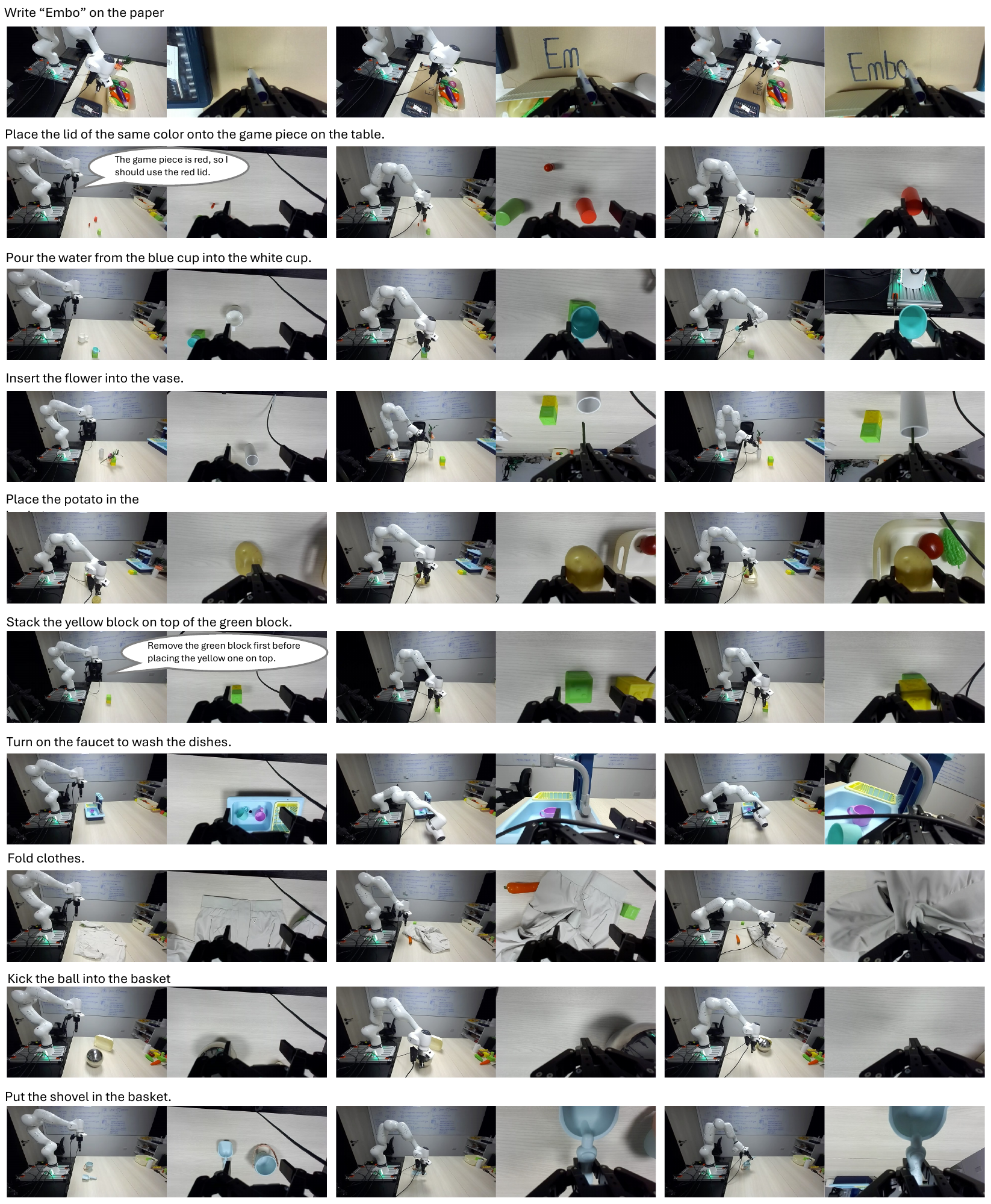}
    \vspace{-2mm}
    \caption{\textbf{Visualization of Real-World Rollout Demos Produced by  Embo-Harness (Zoom in for best view).} On the left is the view from the third-person camera, and on the right is the one from the wrist camera.}
    \label{fig:real_world_demos}
\end{figure*}

\begin{figure*}[t]
    \centering
    \includegraphics[width=\linewidth]{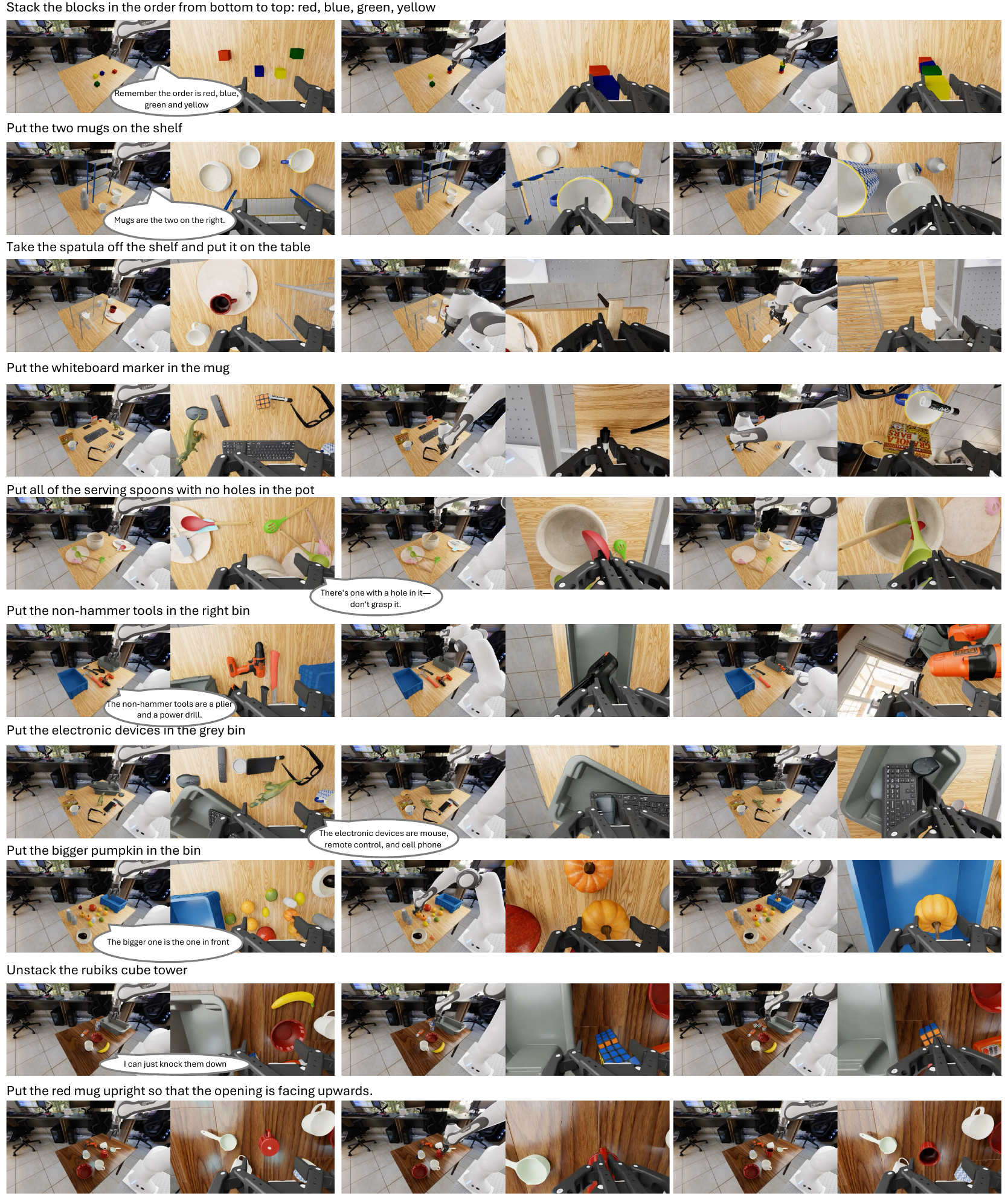}
    \caption{\textbf{Visualization of Simulation Rollout Demos Produced by Embo-Harness (Zoom in for best view).} On the left is the view from the third-person camera, and on the right is the one from the wrist camera.}
    \label{fig:simulation_demos}
\end{figure*}

\end{document}